\documentclass[10pt,journal,compsoc]{IEEEtran}

\usepackage[usenames,svgnames]{xcolor}
\usepackage[ruled,linesnumbered]{algorithm2e} 
\usepackage{graphicx}  
\usepackage{subfig} 
\usepackage{amsfonts}
\usepackage{ dsfont }
\usepackage{hyperref}
\hypersetup{colorlinks,allcolors=black}

\usepackage{ulem}
\usepackage{amssymb}
\usepackage{amsmath}
\usepackage{url}
\usepackage{enumerate}
\usepackage{multirow}
\usepackage{booktabs}
\usepackage{makecell}
\usepackage{bm}
\usepackage{pifont}
\usepackage{mydef}
\usepackage{threeparttable} 
\usepackage{ragged2e}
\usepackage[numbers]{natbib}
\usepackage{siunitx}
\usepackage[edges]{forest}

\newcommand{\cmark}{\ding{51}}
\newcommand{\xmark}{\ding{55}}

\newcommand{\figref}[1]{Figure~\ref{#1}}
\newcommand{\tabref}[1]{Table~\ref{#1}}
\newcommand{\secref}[1]{Section~\ref{#1}}
\renewcommand{\justify}{\leftskip=0pt \rightskip=0pt plus 0cm}

\begin{document}

\title{Graph Prompt Learning: A Comprehensive Survey and Beyond}

\author{Xiangguo~Sun,
        Jiawen Zhang,
        Xixi Wu,
        Hong~Cheng,
        Yun Xiong,
         Jia Li

\IEEEcompsocitemizethanks{\IEEEcompsocthanksitem Xiangguo~Sun, Hong~Cheng: Department of Systems Engineering and Engineering Management, and Shun Hing Institute of Advanced Engineering, The Chinese University of Hong Kong, Hong Kong SAR.  \\\{xgsun, hcheng\}@se.cuhk.edu.hk

\IEEEcompsocthanksitem Jiawen Zhang, Jia Li: Hong Kong University of Science and Technology (Guangzhou), China. \\jzhang302@connect.hkust-gz.edu.cn,jialee@ust.hk

\IEEEcompsocthanksitem Xixi Wu, Yun Xiong: Shanghai Key Laboratory of Data Science, School of Computer Science, Fudan University, China. \\ 21210240043@m.fudan.edu.cn, yunx@fudan.edu.cn
}
}

\markboth{IEEE TRANSACTIONS ON KNOWLEDGE AND DATA ENGINEERING}%
{Xiangguo Sun \MakeLowercase{\textit{et al.}}: }

\IEEEtitleabstractindextext{
\begin{abstract}
\justify{Artificial General Intelligence (AGI) has revolutionized numerous fields, yet its integration with graph data, a cornerstone in our interconnected world, remains nascent. This paper presents a pioneering survey on the emerging domain of graph prompts in AGI, addressing key challenges and opportunities in harnessing graph data for AGI applications. Despite substantial advancements in AGI across natural language processing and computer vision, the application to graph data is relatively underexplored. This survey critically evaluates the current landscape of AGI in handling graph data, highlighting the distinct challenges in cross-modality, cross-domain, and cross-task applications specific to graphs. Our work is the first to propose a unified framework for understanding graph prompt learning, offering clarity on prompt tokens, token structures, and insertion patterns in the graph domain. We delve into the intrinsic properties of graph prompts, exploring their flexibility, expressiveness, and interplay with existing graph models. A comprehensive taxonomy categorizes over 100 works in this field, aligning them with pre-training tasks across node-level, edge-level, and graph-level objectives. Additionally, we present, ProG, a Python library, and an accompanying website, to support and advance research in graph prompting. The survey culminates in a discussion of current challenges and future directions, offering a roadmap for research in graph prompting within AGI. Through this comprehensive analysis, we aim to catalyze further exploration and practical applications of AGI in graph data, underlining its potential to reshape AGI fields and beyond. 
ProG and the website can be accessed by \url{https://github.com/WxxShirley/Awesome-Graph-Prompt}, and \url{https://github.com/sheldonresearch/ProG}, respectively.
}

\end{abstract}

}

\maketitle

\section{Introduction}\label{sec:intro}

In an era marked by the rapid evolution of Artificial General Intelligence (AGI), there emerged many fantastic applications with AGI techniques such as ChatGPT in Natural Language Processing (NLP) and Midjourney in Computer Vision (CV). AGI has greatly improved our lives, making our work more efficient and freeing us from repetitive tasks to focus on more creative endeavors. However, when it comes to working with graph data, AGI applications are still in their early stages compared with the huge success in NLP \cite{
devlin2019bert, 
brown2020language, liu2023pretrain} and CV areas \cite{
wang2023chatvideo, 
zhang2023videollama}. In our increasingly interconnected world, understanding and extracting valuable insights from graphs is crucial. This places AGI applied to graph data at the forefront of both academic and industrial interest \cite{liu2023graph, zhang2023large, yang2023datacentric}, with the potential to redefine fields like drug design \cite{rong2020selfsupervised, qian2023can} and battery development \cite{wang2023scientific}, etc.

\begin{figure}[!t]
    \centering
    \includegraphics[width=0.45\textwidth]{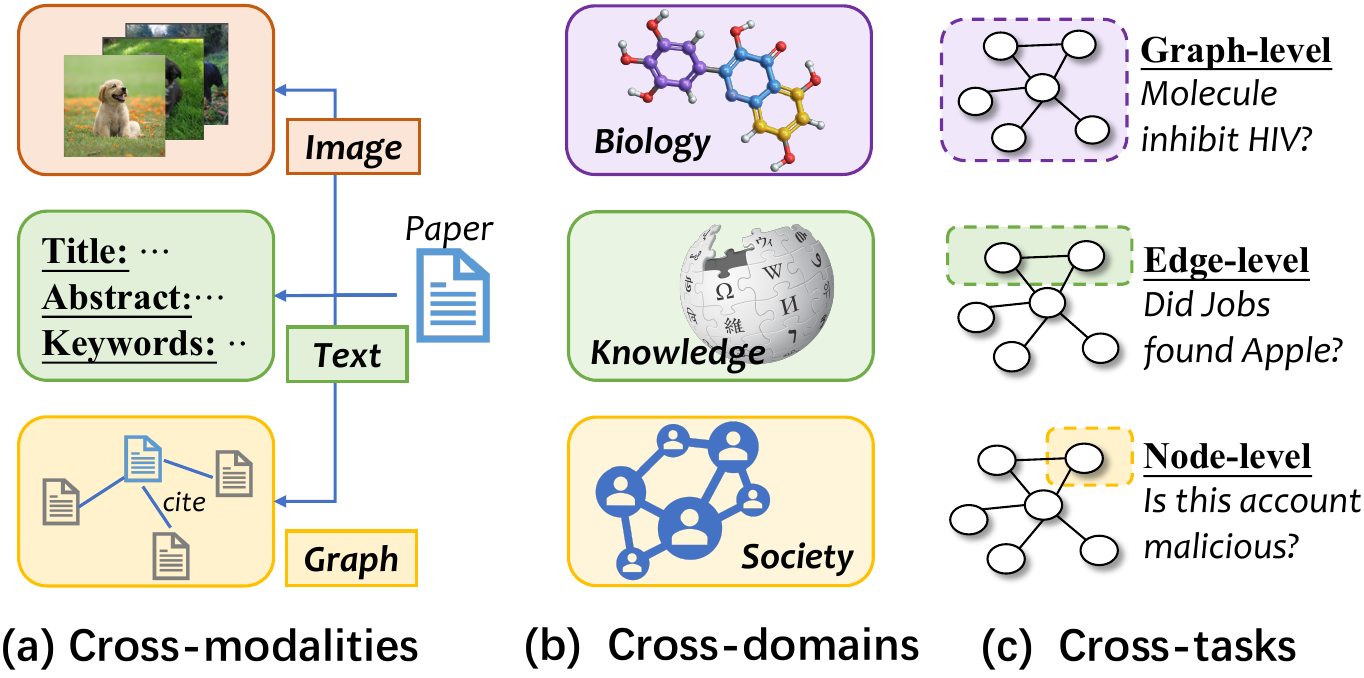} 
    \caption{The Most Frequently Problems towards Artificial General Intelligence. (a) Cross-modalities, (b) Cross-domains, (c) Cross-tasks.}
    \label{fig:AGIProblems}
\end{figure}

However, realizing this vision is never easy. Figure \ref{fig:AGIProblems} illustrates this landscape for recent research in Artificial General Intelligence, from which we can see that there are at least three fundamental problems in technique: How to make the model general for different modalities, different domains, and different tasks? Within the NLP and CV areas, there have been many commercial models that can understand and translate information across these modalities \cite{
devlin2019bert, 
zhang2023videollama, 
brown2020language}. For example, models like BERT \cite{devlin2019bert} and GPT-3 \cite{brown2020language} have demonstrated the ability to perform tasks that involve both textual and visual information. However, in the context of graph data, the harmonization of information from multiple modalities remains largely uncharted territory \cite{li2023graphadapter}.
For the cross-domain issue, transfer learning has proven effective, enabling models to apply knowledge learned from images and text in one domain to another. However, transferring knowledge between different graph domains is very tough because the semantic spaces are not aligned \cite{zhu2023graphcontrol} and the structural patterns are also not similar \cite{ zhao2023graphglow}, making graph domain adaptation remains a very frontier and not well-solved AGI problem. Currently, most graph research on transfer learning focuses on the third problem, how to leverage the pre-trained graph knowledge in the same graph domain to perform different graph tasks (like node classification, link prediction, graph classification, etc) \cite{sun2022gppt, liu2023graphprompt, sun2023all, fang2023universal, zhu2023sglpt, hu2020gptgnn, shirkavand2023deep, ge2023enhancing}. However, compared with the huge success in NLP and CV, task transferring within the same graph domain is still primitive with far fewer instances of successful industrial applications. While the AGI research boasts notable achievements in many linear data like images, 
text \cite{robinson2023leveraging, devlin2019bert, brown2020language}, and videos \cite{wang2023chatvideo, zhang2023videollama}, the fundamental problems within the realm of graph data remain underexplored. Besides the above three foundation problems, Artificial General Intelligence has also encountered many social disputes. For example, training large foundation models consumes exorbitant amounts of energy and may yield unintended counterfactual outcomes \cite{liu2022ptuning, schick2021it}. These concerns have led to a growing consensus within the AI community on the efficient extraction of useful knowledge preserved by these large models, minimizing the need for repetitive fine-tuning across various downstream tasks \cite{gao2021making, lester2021power}. This consensus not only promises to alleviate the environmental impact but also offers a practical solution to the challenge of model efficiency and adaptability in an era of AGI.

\begin{figure}[!t]
    \centering
    \includegraphics[width=0.45\textwidth]{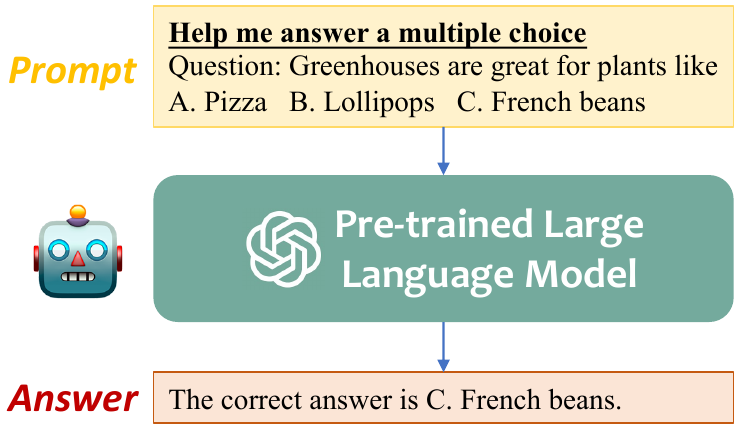} 
    \caption{An example of the language prompt. A textual prompt is designed to let the pre-trained large language model perform the multi-choice question-answering task.}
    \label{fig:PromptExample}
\end{figure}

At the core of recent AGI technology, prompt learning has presented huge potential to solve the above problems and demonstrated remarkable success in NLP and CV applications \cite{qin2021learning, tsimpoukelli2021multimodal, liu2023pretrain}. Prompt learning is the art of designing informative prompts  to manipulate input data for the pre-trained foundation models. Figure \ref{fig:PromptExample} shows an example of a textual-format prompt applied to a pre-trained language model to directly perform downstream inference tasks. By reformulating downstream tasks into pre-training tasks, this approach avoids the need for extensive model tuning and efficiently extracts preserved knowledge \cite{brown2020language, jiang2020how}. Since its powerful capabilities in data manipulation, task reformulation, and extraction of significant insights, prompting is very promising to address the aforementioned cross-modalities, cross-domains, and cross-task challenges in one way. Compared with large models, the prompt is usually very lightweight and can efficiently extract useful knowledge by reducing the extensive computational resources caused by repeat tuning of these large models \cite{lester2021power, shin2020autoprompt}. Intuitively, text and images can be perceived as specific instances of a more general graph data structure. For instance, a sentence can be treated as a graph path, with words as nodes, and an image can be viewed as a grid graph, where each pixel serves as a graph node. This insight encourages us to explore the transference of successful prompting techniques from text to the graph area for similar concerns.

Recently, some researchers have started to introduce prompt learning to graph data \cite{sun2022gppt, liu2023graphprompt, sun2023all, fang2023universal, zhu2023sglpt, ma2023hetgpt, guo2023datacentric, chen2023ultradp, gong2023prompt}. However, some further studies have found that the graph prompt is very different from their counterparts in the NLP area \cite{sun2023all}. First, the design of graph prompts proves to be a far more intricate endeavor compared to the formulation of language prompts. Classic language prompts often comprise predefined phrases or learnable vectors appended to input text \cite{brown2020language, gao2021making}. Here, the primary focus lies in the content of the language prompt. However, we actually do not know what a graph prompt looks like. A graph prompt not only contains the prompt "content" but also includes the undefined task of determining how to structure these prompt tokens and seamlessly integrate them into the original graph. Second, the harmonization of downstream graph problems with the pre-training task is more difficult than language tasks \cite{liu2023graphprompt, sun2023all}. For example, a typical pre-training approach for a language model is to predict a masked word by the model \cite{devlin2019bert}. Then many downstream tasks such as question answering, and sentiment classification can be easily reformulated as word-level tasks \cite{liu2023pretrain}. Unlike NLP, where pre-training tasks often share a substantial task sub-space, graph tasks span node-level \cite{grover2016node2vec}, edge-level \cite{zhang2018link}, and graph-level objectives \cite{sun2020infograph, sun2021heterogeneous}, making pre-training pretexts less adaptable. Third, compared with prompts in NLP which are usually some understandable phrases, graph prompts are usually less intuitive to non-specialists. The fundamental nature and role that graph prompts play within the graph model remain somewhat elusive without comprehensive theoretical analysis. There is also a lack of clear-cut evaluation criteria for the quality of designed graph prompts. In addition, there are still many unclear questions for us to further understand graph prompting. For example, how effective are these graph prompts? What is their efficiency in terms of parameter complexity and training burden? How powerful and flexible do these prompts manipulate the original graph data? In light of these intricacies questions, there is a pressing need for delving deeper into the potential of graph prompts in AGI, thereby paving the way for a more profound understanding of this evolving frontier within the broader data science landscape.

While there have been recent endeavors to explore graph prompting, a consistent framework or clear route remains unavailable. These efforts vary significantly in terms of perspective, methodology, and target tasks, which present a fragmented landscape of graph prompting and pose a considerable obstacle to the systematical advancement of this research area. There arises an urgent need to provide a panoramic view, analysis, and synthesis of the latest advances in this realm with a unified framework. In light of this situation, we offer this survey to present how existing work on graph prompts tries to solve the three foundation problems towards AGI as previously mentioned.  Beyond that, we also wish to push forward the research area by answering the following detailed research questions (RQs): 

\begin{itemize}
    \item \textbf{RQ1: How to understand existing work with a unified framework since they are very different?} Our main focus is to understand the various methods used in the field of graph prompting. We want to bring together all these different approaches and ideas to create a single, cohesive framework. This framework will help us thoroughly grasp the existing research and provide a strong foundation for future research in this area.
   \item  \textbf{RQ2: Why Prompt? What's the Nature of Graph Prompt?}
  In this part of our study, we aim to understand why prompts are important. We want to uncover the fundamental aspects of graph prompts. What exactly do graph prompts do in graph problems? How do they fit into the complex graphs, and how do they help us achieve the broader goal of creating AI systems that can handle graph data effectively? These questions highlight the significant role that graph prompts play in shaping the future of AI when dealing with graph information.

   \item  \textbf{RQ3: How to Design Graph Prompts?} Designing good graph prompts is a complex task. In this part, we explore the technical details of designing graph prompts: what do they look like, how do they align downstream tasks and the pre-train task, and how they are learned? These important questions focus on the skill of making prompts that work well with the complexities of graph data, helping researchers make better prompts.
   

   \item  \textbf{RQ4: How to Deploy Graph Prompts in Real-world Applications?} At the moment, there isn't an easy-to-use toolkit for creating graph prompts. The potential applications that graph prompts can be deployed are under-exploration. This research question focuses on making graph prompts practical for use in real-world situations with an easily extensible programming package.

   \item  \textbf{RQ5: What Are the Current Challenges and Future Directions in Graph Prompting?} This question guides us to look at the challenges we're dealing with today and the way forward. By tackling these important questions, we hope to provide a roadmap for ongoing graph-prompting research.

\end{itemize}

To answer the first research question (RQ1), we propose a unified framework to analyze graph prompt learning work. Our framework casts the concept of a graph prompt into prompt tokens, token structures, and inserting patterns. This higher-level perspective offers clarity and comprehensiveness, providing readers with a structured understanding of this burgeoning field. To the best of our knowledge, our survey marks the first of its kind to bring together these multifaceted aspects of graph prompting within a single unified framework.

To answer the second research question (RQ2), we explore the correlations between prompts and existing graph models through the lenses of flexibility and expressiveness and then present a fresh and insightful perspective to uncover the nature of graph prompts. Unlike most prompt learning surveys in NLP areas \cite{liu2023pretrain} that treat prompting as a trick of filling the gap between the pre-training tasks and downstream tasks, we reveal that graph prompts and graph models are interconnected on a deeper level. This novel perspective offers invaluable insights into why prompt learning holds promise in the graph area and what distinguishes it from traditional fine-tuning methods \cite{hu2020strategies}. To our knowledge, this is the first endeavor to offer such an illuminating perspective on graph prompting.

To answer the third research question (RQ3), we introduce a comprehensive taxonomy that includes more than 100 related works. Our taxonomy dissects these works, categorizing them according to node-level, edge-level, and graph-level tasks, thereby aligning them with the broader context of the pre-training task. This will empower our readers with a clearer comprehension of the mechanisms underlying prompts within the whole "pre-training and prompting" workflow.


To answer the fourth research question (RQ4), we developed ProG (prompt graph) \footnote{\url{https://github.com/sheldonresearch/ProG}}, a unified Python library to support graph prompting. Additionally, we established a website\footnote{\url{https://github.com/WxxShirley/Awesome-Graph-Prompt}} that serves as a repository for the latest graph prompt research. This platform curates a comprehensive collection of research papers, benchmark datasets, and readily accessible code implementations. By providing this accessible ecosystem, we aim to empower researchers and practitioners to advance this burgeoning field more effectively. 

Beyond these, our survey goes a step further with an introduction of potential applications, a thoughtful analysis of the current challenges, and a discussion of future directions, thus providing a comprehensive roadmap for the evolution of this vibrant and evolving field (RQ5). Our contributions are summarised as follows:

\begin{itemize}
    \item \textbf{Enabling Comprehensive Analysis.} We propose a unified framework for analyzing graph prompt learning work, providing a comprehensive view of prompt tokens, token structures, and inserting patterns.     
    \item \textbf{Novel Perspectives on Prompt-Model Interplay. } We offer fresh insights into the nature of graph prompts. Unlike traditional work that simply treats prompts as a trick of filling the gap between downstream tasks and the pre-train task, we explore the flexibility and expressiveness issues of graph models and pioneer a more thorough perspective into the interplay between prompts and existing graph models.
    \item \textbf{A Systematic Taxonomy of Graph Prompting. }We systematically explore over a hundred recent works in the domain of graph prompting. This taxonomy not only organizes these contributions but also furnishes readers with a comprehensive understanding of prompt mechanisms within the overarching "pre-training and prompting" workflow.
    \item \textbf{Empowering the Graph Prompting Ecosystem. }We developed ProG, a Python library supporting graph prompting, and a comprehensive website for collecting the latest graph prompt research.
    \item \textbf{Charting a Path Forward. }A detailed exploration of current challenges and future directions in the field.
\end{itemize}

\textbf{Roadmap.} The rest of this survey is organized as follows: we present our survey methodology in section \ref{sec:method}, followed by preliminaries in section \ref{sec:pre}, the introduction of pre-training methods in section \ref{sec:pretrain_method}, prompting methods for graph models in section \ref{sec:ptask}. We discuss potential applications of graph prompt in section \ref{sec:app} and present our developed library ProG in section \ref{sec:prog}. In section \ref{sec:dis}, we summarize our survey with current challenges and future directions. Section \ref{sec:con} concludes the survey and presents the contribution declaration of the authors.

\section{Survey Methodology}\label{sec:method}

\subsection{Research Objectives}

This survey will introduce the art of prompting from a big picture of artificial general intelligence (AGI). We first present three fundamental problems towards AGI in Table \ref{tab:rqs}. Recently, prompt learning has been demonstrated as a promising solution to these problems in many linear data such as text \cite{brown2020language, gao2021making}, images \cite{
wang2023chatvideo}, etc. However, whether the prompt technique can still solve these problems in the graph area, is not clearly discussed. Through this survey, we wish to figure out how the graph prompt potentially helps graph models to be more general across various tasks and domains, and how it generalizes the foundation models to interact with other modalities (e.g. text, image, etc). Beyond the above common problems of AGI in NLP, CV, and graph areas, graph prompting is usually very different from its counterparts in NLP and CV areas, leading to many detailed questions as shown in Table \ref{tab:rqs}.

\tikzstyle{leaf}=[draw=hiddendraw,
    rounded corners, minimum height=1em,
    fill=mygreen!40,text opacity=1, 
    fill opacity=.5,  text=black,align=left,font=\scriptsize,
    inner xsep=3pt,
    inner ysep=1pt,
    ]
\tikzstyle{middle}=[draw=hiddendraw,
    rounded corners, minimum height=1em,
    fill=output-white!40,text opacity=1, 
    fill opacity=.5,  text=black,align=center,font=\scriptsize,
    inner xsep=7pt,
    inner ysep=1pt,
    ]
\begin{figure*}[ht]
\centering
\begin{forest}
  for tree={
  forked edges,
  grow=east,
  reversed=true,
  anchor=base west,
  parent anchor=east,
  child anchor=west,
  base=middle,
  font=\scriptsize,
  rectangle,
  line width=0.7pt,
  draw=output-black,
  rounded corners,align=left,
  minimum width=2em, s sep=6pt, l sep=8pt,
  },
  where level=1{text width=0.2\linewidth}{},
  where level=2{text width=0.2\linewidth,font=\scriptsize}{},
  where level=3{font=\scriptsize}{},
  where level=4{font=\scriptsize}{},
  where level=5{font=\scriptsize}{},
  [Taxonomy, middle,rotate=90,anchor=north,edge=output-black
      [Pre-training Strategies for \\Graph Prompt (\secref{sec:pretrain_method}),middle,anchor=west,edge=output-black, text width=0.18\linewidth
        [{Node-level Strategies\\(\secref{subsec:node_pre})}, middle, text width=0.16\linewidth, edge=output-black
             [\citet{cheng2023wiener, zhu2021graph, jin2021multiscale, peng2020graph}\\\citet{hamilton2017inductive,hou2022graphmae,wang2021selfsuperviseda,wang2017mgae,park2019symmetric}\\\citet{hou2023graphmae2,wang2021selfsuperviseda,jiang2021pretraining}, leaf, text width=0.46\linewidth, edge=output-black] ]
        [{Edge-level Strategies\\(\secref{subsec:edge_pre})}, middle, text width=0.16\linewidth, edge=output-black
             [\citet{jin2021node, long2022pretraining,tan2023s2gae, li2023what,pan2018adversarially}\\\citet{ hasanzadeh2019semiimplicit, kim2021how}, leaf, text width=0.46\linewidth, edge=output-black] ]
        [{Graph-level Strategies\\(\secref{subsec:graph_pre})}, middle, text width=0.16\linewidth, edge=output-black
             [\citet{xie2022selfsupervised, rong2020selfsupervised,velickovic2019deep, sun2020infograph}\\\citet{sun2021mocl, subramonian2021motifdriven,you2020graph,suresh2021adversarial}\\\citet{ thakoor2021bootstrapped,qiu2020gcc}, leaf, text width=0.46\linewidth, edge=output-black]]
        [{Multi-task Pre-training\\(\secref{subsec:multi_pre})}, middle, text width=0.16\linewidth, edge=output-black
             [\citet{hu2020strategies, hu2020gptgnn, zhang2020graphbert,fang2022geometryenhanced}, leaf, text width=0.46\linewidth, edge=output-black] ]]
    [{Prompt Design for \\Graph Tasks (\secref{sec:ptask})}, middle,anchor=west,edge=output-black, text width=0.18\linewidth
        [{Prompt Token, Structure, and \\Inserting Pattern (\secref{subsec:pdesign})}, middle, text width=0.18\linewidth, edge=output-black
            [Prompt as Tokens, middle, text width=0.1\linewidth, edge=output-black [\citet{zhu2023graphcontrol,fang2022prompt,fang2023universal}\\\citet{gong2023prompt,tan2023virtual,liu2023graphprompt}\\\citet{zhu2023sglpt,ma2023hetgpt,sun2022gppt}\\\citet{chen2023ultradp,shirkavand2023deep}, leaf, text width=0.3\linewidth, edge=output-black]]
            [Prompt as Graphs, middle, text width=0.1\linewidth, edge=output-black
                [\citet{sun2023all,huang2023prodigy,ge2023enhancing}, leaf, text width=0.3\linewidth, edge=output-black] ] ]
        [Aligning Tasks by Answering \\Function (\secref{subsec:align}), middle, text width=0.18\linewidth, edge=output-black
             [Handling Different Level Tasks, middle, text width=0.18\linewidth, edge=output-black
             [\citet{sun2023all,huang2023prodigy}\\\citet{liu2023graphprompt,gong2023prompt}, leaf, text width=0.22\linewidth, edge=output-black] ]
              [Learnable Answering Function, middle, text width=0.18\linewidth, edge=output-black
                [\citet{sun2023all,fang2022prompt}\\\citet{fang2023universal,tan2023virtual}, leaf, text width=0.22\linewidth, edge=output-black]]
              [Hand-crafted Answering Function, middle, text width=0.2\linewidth, edge=output-black [\citet{sun2023all,sun2022gppt}\\\citet{ma2023hetgpt,ge2023enhancing}\\\citet{huang2023prodigy,liu2023graphprompt}, leaf, text width=0.2\linewidth, edge=output-black ]]]
        [{Prompt Tuning (\secref{subsec:pt})}, middle, text width=0.18\linewidth, edge=output-black
             [Meta-Learning Technique, middle, text width=0.2\linewidth, edge=output-black
                [\citet{sun2023all}, leaf, text width=0.2\linewidth, edge=output-black ] ]
             [Task-specific Tuning, middle, text width=0.2\linewidth, edge=output-black
                [\citet{fang2022prompt}, leaf, text width=0.2\linewidth, edge=output-black] ]
             [Tuning in Line with Pretext, middle, text width=0.2\linewidth, edge=output-black [\citet{liu2023graphprompt,sun2022gppt}\\\citet{tan2023virtual,ma2023hetgpt}\\\citet{ge2023enhancing,gong2023prompt}\\\citet{chen2023ultradp}, leaf, text width=0.2\linewidth, edge=output-black] ] ]
        [{Connection, Pros, and Cons\\(\secref{subsec:pdiscussion})}, middle, text width=0.18\linewidth, edge=output-black [\citet{fang2022prompt,fang2023universal,sun2023all,ma2023hetgpt,liu2023graphprompt}\\\citet{sun2022gppt,huang2023prodigy,chen2023ultradp,tan2023virtual}, leaf, text width=0.44\linewidth, edge=output-black] ] ]
    [Multi-Modal Prompting \\with Graphs (\secref{subsec:multi_modal}),middle, anchor=west, edge=output-black, text width=0.18\linewidth
         [Prompt in Text-Attributed Graphs, middle, text width=0.2\linewidth, edge=output-black
            [\citet{wen2023augmenting,wen2023prompt,zhao2023gimlet,li2023promptbased}, leaf, text width=0.42\linewidth, edge=output-black] ]
         [Large Language Models in Graph \\Data Processing, middle, text width=0.2\linewidth, edge=output-black
           [\citet{chen2023exploring,fatemi2023talk,jin2023patton,wang2023knowledge}, leaf, text width=0.42\linewidth, edge=output-black ] ]
         [Multi-modal Fusion with Graph \\and Prompting, middle, text width=0.2\linewidth, edge=output-black
           [\citet{edwards2023synergpt,li2023graphadapter,liu2023gitmol}, leaf, text width=0.42\linewidth, edge=output-black ] ]  ]
    [Graph Domain Adaptation\\ with Prompting (\secref{subsec:pdomain}),middle, anchor=west, edge=output-black, text width=0.18\linewidth
         [Semantic Alignment, middle, text width=0.2\linewidth, edge=output-black [\citet{sun2023all,zhu2023graphcontrol,zhang2023structure,yi2023contrastive}\\\citet{liu2023one}, leaf, text width=0.42\linewidth, edge=output-black ]
        ]
        [Structural Alignment, middle, text width=0.2\linewidth, edge=output-black
    [\citet{shirkavand2023deep,cao2023when,zhao2023graphglow}\\\citet{guo2023datacentric}, leaf, text width=0.42\linewidth, edge=output-black ] ] ]
]
\end{forest} 
\caption{Taxonomy of this Survey with Representative Works.}
\label{fig:Taxonomy}
\end{figure*}
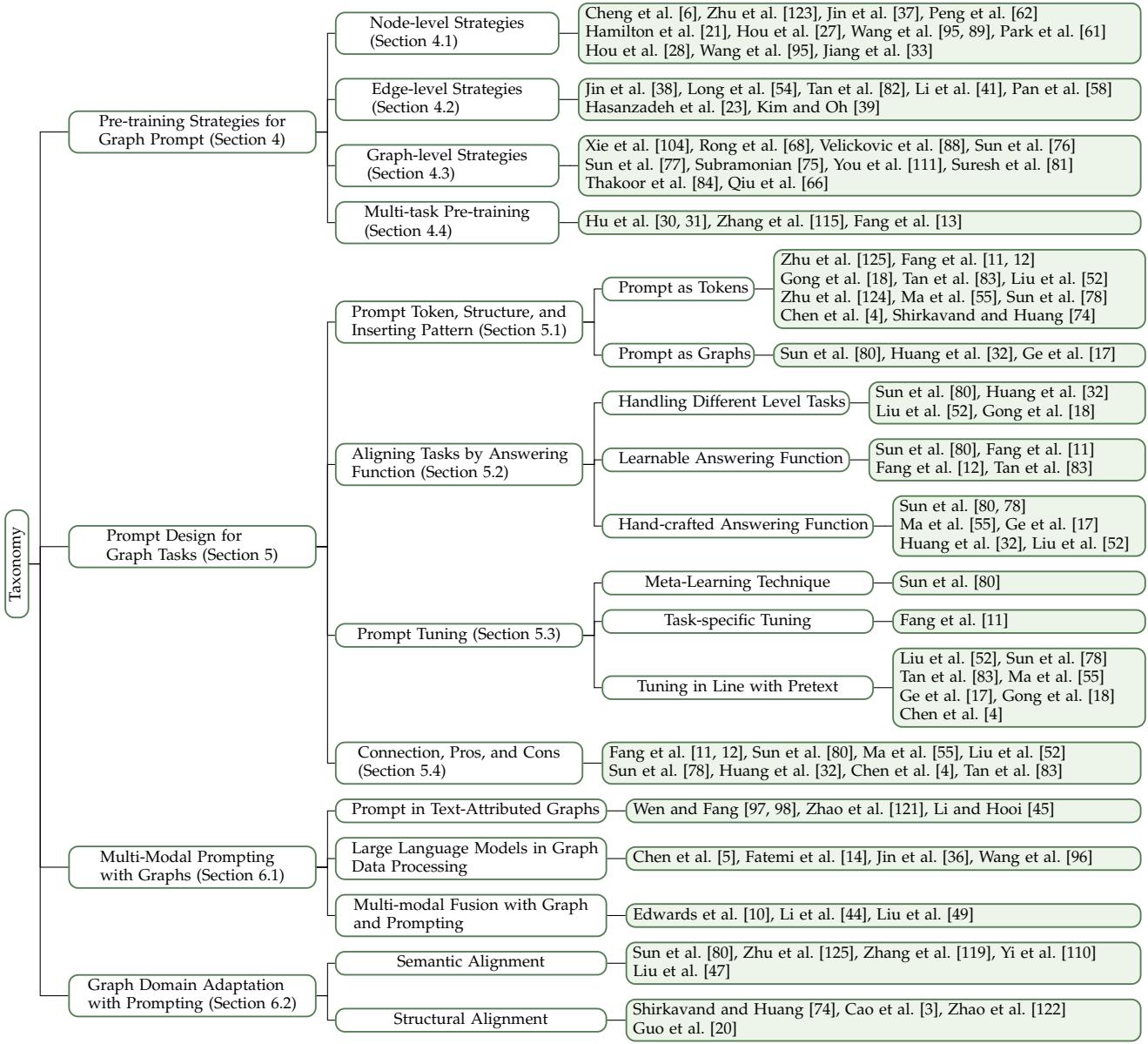

\subsection{Taxonomy}

The taxonomy of this survey is presented in \figref{fig:Taxonomy}, which is intricately designed to categorize graph prompts based on their specific applications and functionalities, providing a structured approach to understanding their role in AGI.

\textbf{(1) Pre-training Strategies for Graph Prompt.}

Since prompting techniques mostly seek to reformulate downstream tasks to the pre-training tasks, they are highly customized for detailed pre-training approaches, thus we briefly discuss the representative pre-training work in the graph area before we formally introduce graph prompt content. We split existing pre-training approaches into node-level, edge-level, graph-level, and multi-task pre-training strategies. Next, we present how different prompting ideas reformulate various downstream tasks to the corresponding pre-training tasks in the graph area. By revisiting existing pre-training literature, we will be more clear about the role of graph prompts in the whole "pre-training and prompting" framework.

\textbf{(2) Prompting Methods in Graph Areas.} 

Aiming at the three foundation problems mentioned in the research objectives (P1-P3 in \tabref{tab:rqs}), we analyze graph prompting from three aspects: \romannumeral1. prompt design for graph tasks (\secref{sec:ptask}); \romannumeral2. multi-modal prompting with graphs (\secref{subsec:pmodal}); and \romannumeral3. graph domain adaptation with prompting techniques (\secref{subsec:pdomain}). Within each aspect, we present a detailed discussion related to the five specific problems in the graph prompt area (Q1-Q5 in \tabref{tab:rqs}).

\textit{\romannumeral1. Prompt Design for Graph Tasks.} In this section, we propose a unified framework to analyze existing works on graph prompt design. Our framework treats existing graph prompts with three key components: \textit{prompt tokens}, which preserve prompt content as vectors; \textit{token structures}, which indicate how multiple tokens are organized; and \textit{inserting patterns}, which define how to combine graph prompt with the original graphs. Beyond that, we also carefully analyze how these works design the prompt answering function, which means how they get results for the downstream tasks from their  prompts. We also summarize three representative methods to learn appropriate prompts, including meta-learning techniques, task-specific tuning, and tuning in line with pretext. In the end, we further discuss these works in \secref{subsec:pdiscussion} to see their intrinsic connections with pros and cons.

\textit{\romannumeral2. Multi-modal Prompting with Graphs.} In this section, we briefly present how graph prompts work in a text-attributed graph, which can be seen as the fusion of text and graph modalities. With the progress of large language models (LLMs), the fusion of text and graph data has become easier and has aroused a lot of work on this topic. Since the topic of integrating LLMs with graphs has been well summarized in \cite{li2023survey}, we won't present too much in this survey. Instead, we only briefly discuss some representative works in this area that focus on the prompt area.

\begin{table}[!t]
\centering
\caption{Research Objectives}
\label{tab:rqs}
\begin{tabular}{@{}l|l@{}}
\toprule
\multirow{3}{*}{\makecell[c]{Foundamental \\Problems\\ towards AGI}}  
        & \makecell[l]{P1: How to Make the Model General for Different\\ Modalities? (\secref{subsec:multi_modal}) } \\\cmidrule(l){2-2} 
        & \makecell[l]{P2: How to Make the Model General for Different \\Domains? (\secref{subsec:pdomain})} \\\cmidrule(l){2-2} 
        & \makecell[l]{P3: How to Make the Model General for Different \\Tasks?  (\secref{sec:ptask}) }  \\ \midrule
\multirow{5}{*}{\makecell[c]{Detailed \\Questions of\\ Graph Prompt}} 
        & \makecell[l]{Q1: How to Understand Existing Work with a \\Unified Framework? (\secref{subsec:pdesign})} \\\cmidrule(l){2-2} 
        & \makecell[l]{Q2: What's the Nature of Graph Prompt? \\(\secref{subsec:why_prompt}, \secref{subsec:pdiscussion})}   \\\cmidrule(l){2-2} 
        & \makecell[l]{Q3: How to Design Graph Prompts? (\secref{sec:ptask}) } \\\cmidrule(l){2-2} 
        & \makecell[l]{Q4: How to Deploy Graph Prompts in Real-world \\Applications?  (\secref{sec:app}, \secref{sec:prog})}  \\\cmidrule(l){2-2} 
        & \makecell[l]{Q5: What Are the Current Challenges and Future \\Directions? (\secref{sec:challenges}) }   \\ \bottomrule
\end{tabular}%
\end{table}

\textit{\romannumeral3. Graph Domain Adaptation through Prompting Techniques.} In this section, we introduce related work from two branches. The first branch presents works solving semantic alignment across different graph  domains, and the second branch presents structural alignment. 

\subsection{Literature Overview}
\begin{figure*}[!t]
\centering
\subfloat[Venue distribution of collected papers.]{
\label{fig:venue_dis}
\includegraphics[width=0.4\textwidth]{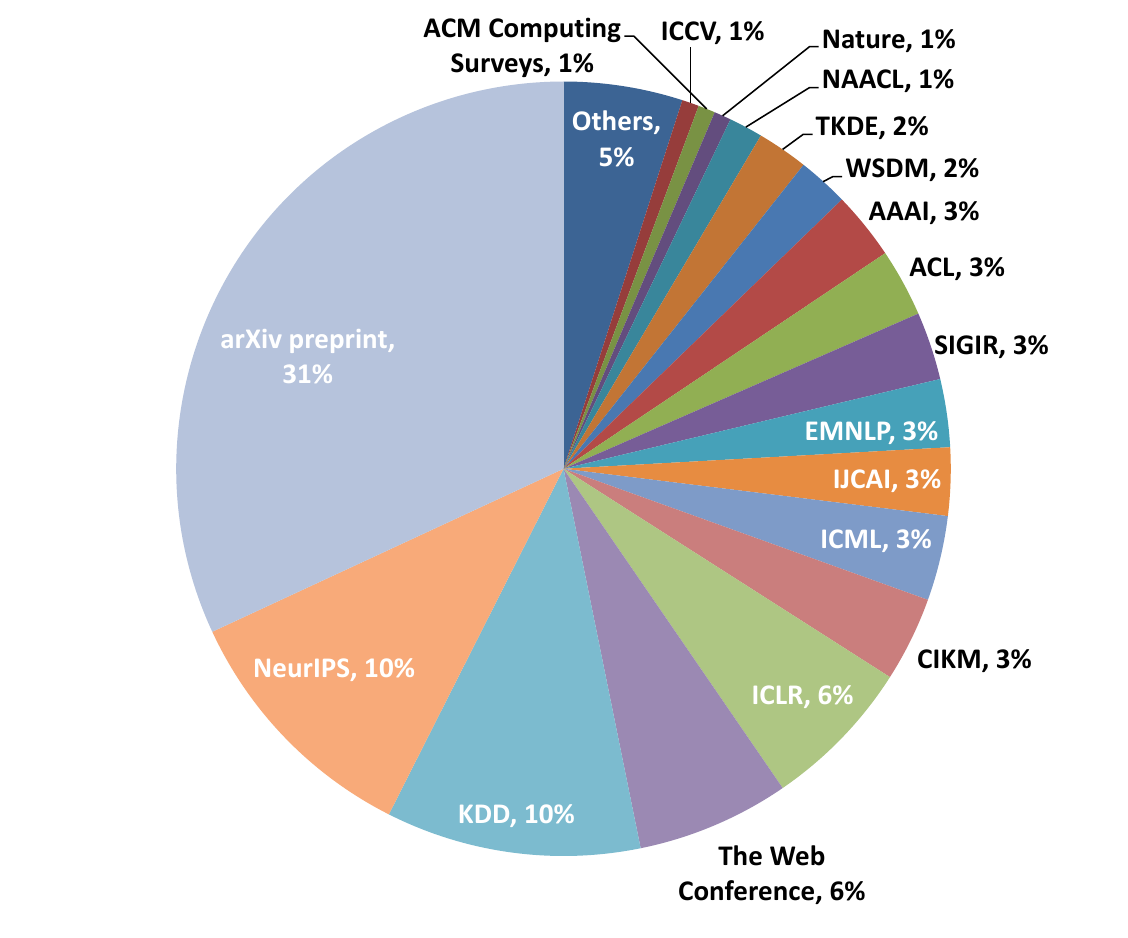}%
} 
\subfloat[Top 15 Keywords appeared in titles of collected papers]{
\label{fig:top10keys}
\includegraphics[width=0.5\textwidth]{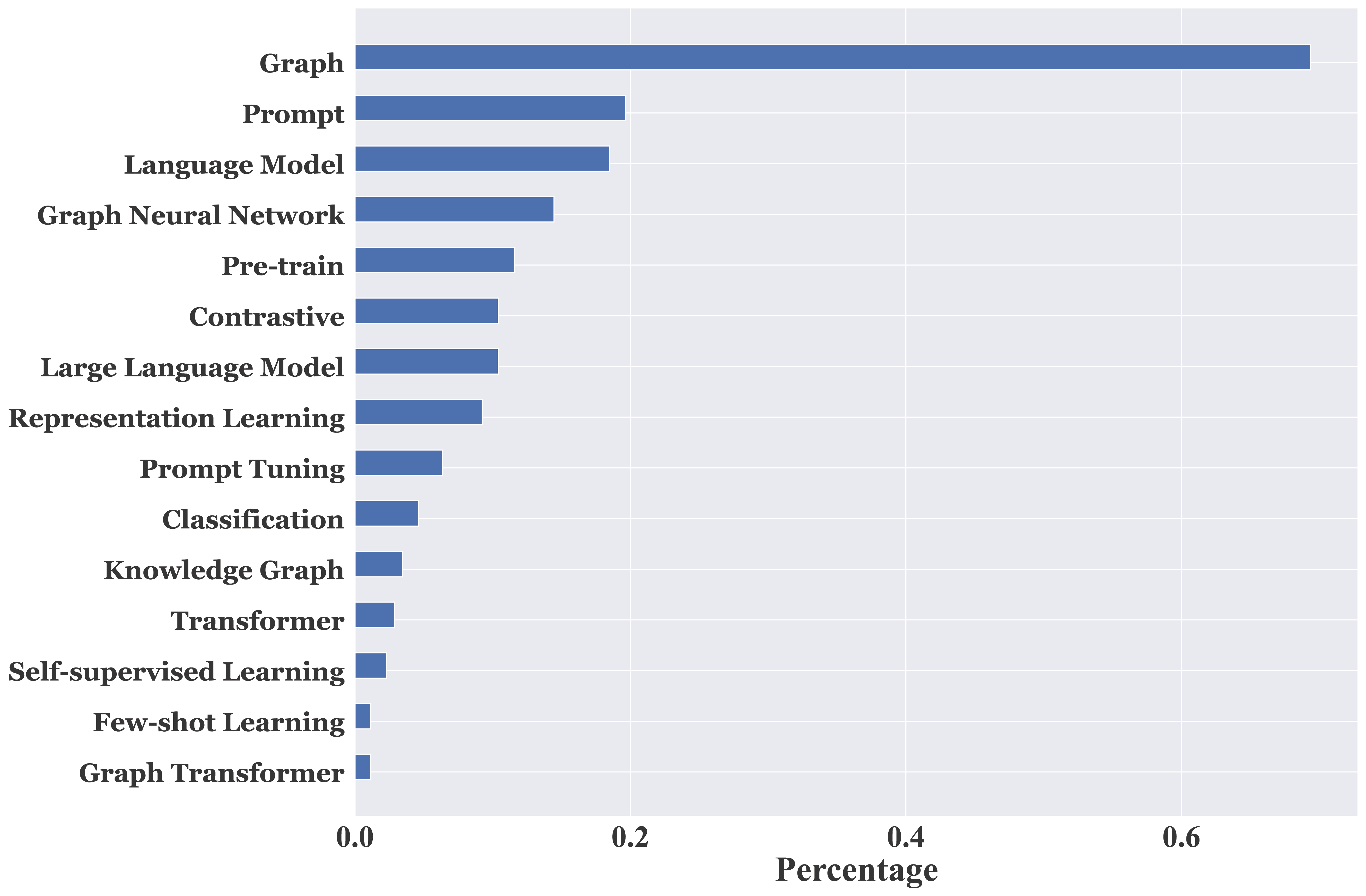}%
}
\caption{Statistics of the collected papers}\vspace{-3ex}
\label{fig:credit_case}
\end{figure*}

In this survey, we carefully studied more than 100 high-quality papers published within the past 5 years from reputable conferences and journals including but not limited to NeurIPS, SIGKDD, The Web Conference, ICLR, CIKM, ICML, IJCAI, EMNLP, SIGIR, ACL, AAAI, WSDM, TKDE, etc. Most of these venues are ranked as CCF A\footnote{\url{https://www.ccf.org.cn/Academic_Evaluation/By_category}} or CORA A*\footnote{\url{https://www.core.edu.au/conference-portal}}. Besides these works, we also introduce several latest works in arXiv so that our survey can catch up with the frontier and latest progress in this area. A more detailed pie chart (Figure \ref{fig:venue_dis}) presents the distribution of collected papers over these venues. Furthermore, we conducted an analysis of the topics covered by these references. In Figure \ref{fig:top10keys}, we present the top 15 keywords that appeared in the titles of these papers. Notably, these keywords align closely with the focus of our survey, which is centered around graph domains and prompt learning.

\noindent \textbf{Connection to Existing Work: }

Our survey stands out from existing surveys in several notable ways. For example, \citet{liu2023graph} primarily focuses on graph foundation models (GFMs). Their survey does not specifically target graph prompts, and only a few papers in this area are briefly discussed. \citet{li2023survey} systematically analyze recent works that integrate graphs and LLMs, which is a detailed analysis of a small portion (Section \ref{subsec:multi_modal}) in our survey. We go beyond their scope by exploring various aspects of graph prompts in a more extensive manner. Meanwhile, surveys \cite{xie2023selfsupervised, yu2023selfsupervised} focus primarily on the pre-training stages, without involving the crucial aspect of graph prompt learning. While a prior survey on graph prompt learning by \citet{wu2023survey} exists, our survey surpasses it in several key aspects. Firstly, we provide \textbf{a more comprehensive analysis of related works}. Their survey was published in May 2023 when there were only a few graph prompt works available \cite{sun2022gppt, zhu2023sglpt, fang2022prompt}. In contrast, our survey encompasses a broader scope, including all relevant works in the field. Secondly, we offer \textbf{a systematic analysis of existing works} within a uniform framework, facilitating comprehension and comparison between different approaches. Thirdly, our survey provides deep insights into \textbf{the relationship between graph pre-training and prompts}, shedding light on the interplay between these critical elements. Lastly, we not only present empirical insights but also include \textbf{engineering works} aimed at deploying graph prompts in real-world applications, ensuring the practical applicability of our survey. 

\section{Preliminaries}\label{sec:pre}

\begin{figure}[!t]
    \centering
    \includegraphics[width=0.49\textwidth]{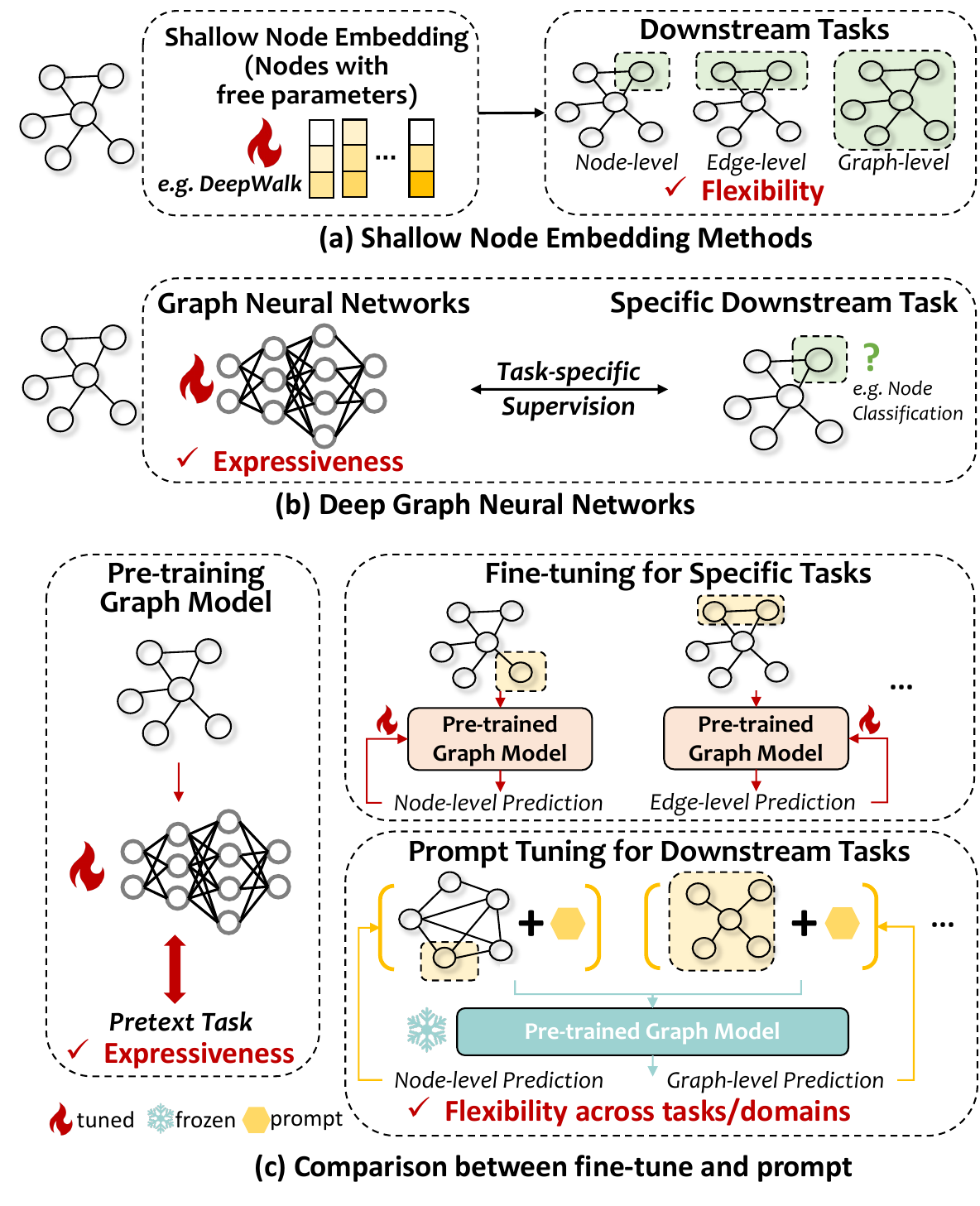}
    \caption{Our perspective of prompt upon flexibility and expressiveness. (a) Shallow node embedding methods offer flexibility across different downstream tasks but sacrifice expressiveness. (b) GNNs provide expressiveness but require task-specific supervision and constant node features, limiting their flexibility. (c) Prompts enable a balanced approach, achieving both flexibility and expressiveness.}
    \label{fig:pre1}
\end{figure}

Graph representation learning has been a topic of extensive research over the past few decades. This journey, illustrated in Figure~\ref{fig:pre1}, has witnessed the evolution from shallow embedding methods to supervised graph neural networks, transitioning from the fine-tuning paradigm to the emerging prompting paradigm. In this section, we will provide an overview of the fundamental notations employed in this survey, delve into the historical developments of graph representation learning, explore the pre-training and fine-tuning paradigm, and trace the evolution of prompt-based learning. Most importantly, we will present a novel perspective focusing on flexibility and expressiveness, shedding light on why prompts offer a promising solution to address the limitations of existing graph representation learning methods.

\subsection{Notations}\label{subsec:notation}
Let a graph instance denoted as $\mathcal{G} = \{ \mathcal{V}, \mathcal{E} \}$, where $\mathcal{V} = \{ v_1,v_2, \ldots, v_N \}$ represents the node set containing $N$ nodes. The edge set $\mathcal{E} \in \mathcal{V} \times \mathcal{V}$ describes the connection between nodes. Each node $v_i$ is associated with a feature vector represented as $\textbf{x}_i \in \mathbb{R}^D$. To characterize the connectivity within the graph, we employ the adjacency matrix denoted as $\textbf{A} \in \{0,1\}^{N \times N}$, where the entry $\textbf{A}_{ij} = 1$ if and only if the edge $(v_i,v_j) \in \mathcal{E}$.

\subsection{Graph Representation Learning}
The last decades have witnessed a notable surge in the development of graph representation learning techniques. These approaches can be broadly categorized into two main branches: shallow embedding methods and deep graph neural networks (GNNs). The shallow embedding approach is centered on mapping nodes into lower-dimensional, learnable embeddings, enhancing their applicability in various downstream tasks, as exemplified by node2vec~\cite{grover2016node2vec} and DeepWalk~\cite{perozzi2014deepwalk}. On the other hand, the deep GNNs maintain the input node features as constants and optimize deep graph model parameters for specific tasks, leading to more expressive representation capabilities, as seen in methods such as Graph Convolution Networks (GCN)~\cite{niepert2016learning} and GraphSAGE~\cite{hamilton2017inductive}.

Shallow embedding approaches make input node features learnable parameters, aiming at encoding nodes in a manner that retains the original network's similarity structure. 
According to node similarity definition, these methods can be categorized as factorization-based~\cite{
ou2016asymmetric, zhang2019prone} and random walk approaches~\cite{grover2016node2vec, perozzi2014deepwalk, wang2019hyperbolic, wang2022common}. 
Despite the flexibility that shallow embedding methods offer for various downstream tasks, they are constrained by their inability to generate embeddings for nodes not encountered during training. Additionally, these approaches lack the capability to incorporate node features. Therefore, more ``deeper" methods, specifically those based on graph neural networks, have been developed to address these limitations.

Most deep GNNs follow a message-passing schema and use a more complex encoder, resulting in powerful expressiveness in graph representation. The representative neural network structure is convolutional graph neural networks (ConvGNNs), which comprise spectral~\cite{defferrard2016convolutional,he2021bernnet}
and spatial methods~\cite{
niepert2016learning, hamilton2017inductive, 
velickovic2018graph, shi2021masked, xu2018how}.
While these methods have exhibited remarkable capabilities in various graph-based applications, their reliance on task-specific supervision imposes constraints on their adaptability and generalizability, particularly when dealing with tasks that have limited labeled data.

In summary, shallow embedding methods offer flexibility, preserving network structure and node content for straightforward graph analytic tasks. However, they lack expressiveness and the ability to encapsulate additional node features. Conversely, GNNs provide more expressive graph representations but require task-specific training data, limiting their transferability. Hence, it calls for a graph learning mechanism that combines expressiveness and flexibility. This need led to the development of the pre-training and fine-tuning paradigm.

\subsection{Pre-training and Fine-tuning}
To address the challenges of limited labeled data and generalization issues in GNNs, the pre-training and fine-tuning paradigm, thriving in the natural language processing (NLP) community, has gained widespread adoption in graph representation learning. These approaches involve pre-training models on large-scale graph data, with or without labels, followed by fine-tuning model parameters for diverse downstream tasks. This two-step process improves model initialization, yielding broader optima and enhanced generalization compared to training from scratch. Commonly employed pre-training schemes include Graph AutoEncoders (GAEs)~\cite{
wang2017mgae}, Masked Components Modeling (MCM)~\cite{hu2020strategies, rong2020selfsupervised}, Graph Contrastive Learning (GCL)~\cite{velickovic2019deep, sun2020infograph}, etc. In Section~\ref{sec:pretrain_method}, we will delve into a detailed discussion of the pre-training and fine-tuning method, offering a comprehensive picture.

\subsection{A Brief History of Prompt Learning}\label{subsec:his_prompt}

Due to the growing number of model parameters, the conventional \emph{pre-training and fine-tuning} process is evolving into a new approach termed \emph{pre-training, prompting, and predicting}~\cite{liu2023pretrain}. In this paradigm, instead of manually adapting the pre-trained model for specific downstream tasks, these tasks are reformulated to resemble those addressed during the pre-training phase, aided by prompts.
Prompts in NLP take various shapes, including cloze prompts
, which complete textual strings like those used in masked language models, and prefix prompts~\cite{lester2021power,li2021prefixtuning}, where the input text precedes the answer slot, as employed by autoregressive language models. Some studies involve manually designed templates based on human insights~\cite{
brown2020language,schick2021fewshot,schick2021it}, while others explore automated template learning. This includes searching for templates in a discrete space~\cite{jiang2020how, haviv2021bertese, shin2020autoprompt,gao2021making} or conducting prompting directly in the embedding space~\cite{li2021prefixtuning, lester2021power,tsimpoukelli2021multimodal,qin2021learning}.
Such a paradigm enables a single pre-trained model to address a multitude of downstream tasks in an unsupervised manner, which has been widely demonstrated by large language models. In light of this, the application of prompting techniques in the context of graph-based tasks is currently an area of active exploration.

\subsection{Why Prompt? A New Perspective upon Flexibility and Expressiveness.}\label{subsec:why_prompt}

Why is prompt learning promising for the graph domain? An existing perspective that appears in most related work is that prompt can reformulate downstream tasks to the pre-training task, which might fill the gap between them. This perspective is good but still not profound enough to see the intrinsic difference from traditional fine-tuning. For example, in a similar perspective, pre-training and fine-tuning can be treated as using fine-tuning to reformulate the pre-training task to the downstream task. It seems that these two technique routines can both address the same problem. Why the first choice is better than the second one?

In this section, we propose a new perspective, from this view, we can further see the difference between prompting and fine-tuning. As discussed in previous sections, existing graph representation learning methods fail to achieve a satisfactory trade-off between expressiveness and flexibility. Shallow graph embedding approaches offer flexibility as they can be applied to a wide range of downstream tasks. However, they sacrifice expressiveness due to limited parameterization and the inability to incorporate original node features. Take the DeepWalk model as an example, shallow graph methods usually treat node representations as free parameters, which is very flexible because each node can learn its individual representations independently. However, for the reason of gradient, they can not rely on more complicated networks later, which might lose some expressiveness. Actually, there are many more advanced works with deep graph layers using the node representations from DeepWalk as their input features as these node representations are very general in various tasks. On the other hand, GNN-based methods treat node embedding as constant features and seek to find a powerful network for mapping node features to a specific task, which is very expressive. However, the learned feature transform pattern is applied to all the nodes, which means the model can not treat each node embedding as free parameters, and can not achieve as flexible results as the previous ones. When we have multiple tasks, we usually need to train different versions of the same GNN model, which is not as flexible as the previous one.

With the above analysis, we can find that traditional fine-tuning actually seeks to further improve the expressiveness of a new task with the pre-trained graph model and can not take care of node flexibility. Unlike fine-tuning, a graph prompt usually has several tokens with free parameters, which is very similar to shallow graph methods. In the meantime, each node in the original graph has constant features for GNN models. By inserting the prompt graph to the original graph, the combined graph has both nodes with constant features and tokens with free parameters. The token parameters can be efficiently tuned, preserving node flexibility. The combined graph is sent to a frozen pre-trained GNN model to leverage the powerful expressiveness of deep graph models.

In this paper, we argue that the prompting mechanism offers a promising solution to address the limitations of existing graph representation learning methods, effectively balancing flexibility and expressiveness. Pre-trained GNNs inherently possess knowledge of both structural and semantic aspects, enabling the desired level of expressiveness. By introducing prompts, we can seamlessly apply powerful pre-trained models to diverse downstream tasks across various domains in an efficient manner. This is achieved by aligning the format of downstream tasks with that of pre-trained tasks, thus leveraging the full potential of pre-trained models even with minimal supervision signals. While the fine-tuning mechanism can also facilitate domain or task adaptation of pre-trained graph models, it often necessitates a considerable amount of labeled information and requires exhaustive re-training of the pre-trained model. In comparison, the prompt mechanism offers a higher degree of flexibility and efficiency.

\section{Pre-training GNNs for Graph Prompting}\label{sec:pretrain_method}

\begin{figure}[t]
    \centering
    \includegraphics[width=0.49\textwidth]{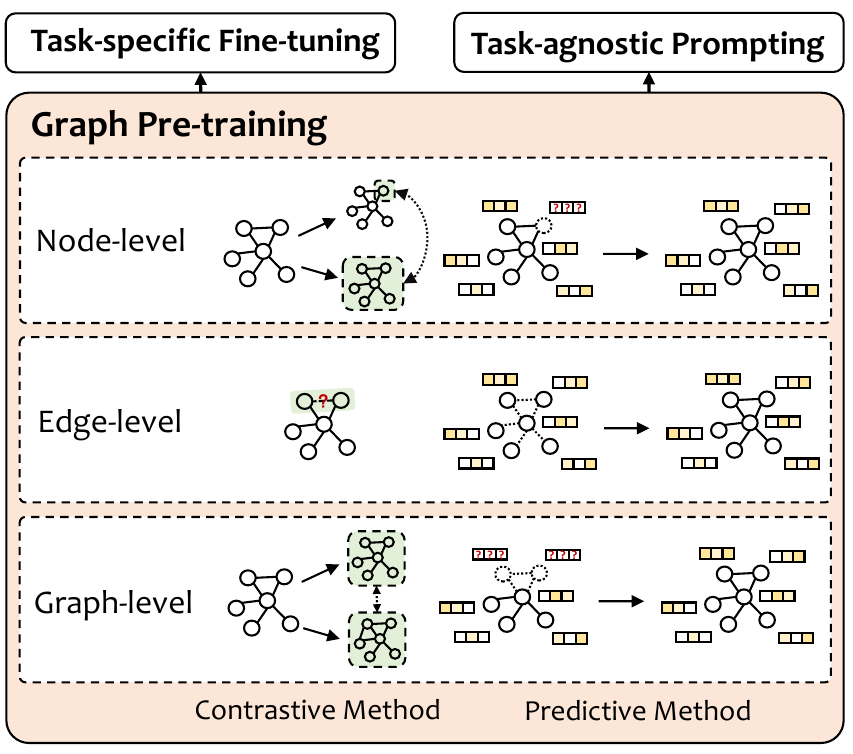}
    \caption{Graph pre-training methods.}
    \label{fig:pretrain}
\end{figure}

Graph pre-training is a pivotal step of the pre-training, prompting, and predicting paradigm in graph representation learning. This approach leverages readily available information to encode the inherent graph structure, providing a robust foundation for generalization across diverse downstream tasks. By integrating these pre-training methods into the comprehensive workflow, we offer an exploration of their interplay with the subsequent prompting and predicting phases, shedding light on the strengths and limitations of this holistic approach. 
This unique perspective distinguishes our survey, framing graph pre-training as an integral part of the broader graph-prompting learning process.
To better illustrate the motivation behind the prompting paradigm, we will now delve into four distinct pre-training strategies within the traditional pre-training and fine-tuning framework.

\subsection{Node-level Strategies}\label{subsec:node_pre}
Node-level pre-training strategies empower the acquisition of valuable local structure representations that can be transferred to downstream tasks. As shown in Figure~\ref{fig:pretrain}, these strategies encompass both contrastive and predictive learning methods. In contrastive learning, self-supervised signals typically result from perturbations in the original graph structure or attributes, with the goal of maximizing Mutual Information (MI) between the original and self-supervised views. Noteworthy node-level contrastive methods include those presented in~\cite{cheng2023wiener, zhu2021graph, jin2021multiscale, peng2020graph, jiang2021pretraining, wang2021selfsuperviseda}. On the other hand, predictive models focus on reconstructing perturbed data using information from the unperturbed data, as demonstrated in~\cite{hamilton2017inductive, hou2022graphmae, wang2017mgae,park2019symmetric,wang2021selfsupervised,hou2023graphmae2}. However, its emphasis on partially semantic topology patterns restricts its ability to capture higher-order information.

\subsection{Edge-level Strategies}\label{subsec:edge_pre}
To enhance performance in tasks such as link prediction, diverse edge-level pre-training strategies have been developed. These strategies excel at capturing node interactions and have undergone extensive exploration. One approach involves discriminating the presence of edges between pairs of nodes, which can be regarded as contrastive methods~\cite{jin2021node, long2022pretraining}. Another approach focuses on reconstructing masked edges by recovering the adjacency matrix~\cite{tan2023s2gae, li2023what, pan2018adversarially, hasanzadeh2019semiimplicit, kim2021how}. Although this pre-training strategy performs admirably in tasks closely related to predicting node relations, it concentrates solely on structural aspects, neglecting the portrayal of node properties, and may encounter challenges when applied to graph-level downstream tasks.

\subsection{Graph-level Strategies}\label{subsec:graph_pre}
The necessity to improve graph-level representations for subgraph-related downstream tasks has prompted the exploration of various graph-level pre-training strategies. Similar to node- and edge-level strategies, these approaches can be broadly categorized into two main groups: graph reconstruction methods, involving the masking of graph components and their subsequent recovery~\cite{xie2022selfsupervised, rong2020selfsupervised}, and contrastive methods focused on maximizing mutual information. These contrastive methods target either local patches of node features and global graph features~\cite{velickovic2019deep, sun2020infograph, hassani2020contrastive, sun2021mocl, subramonian2021motifdriven}, or positive and negative pairs of graphs~\cite{you2020graph, suresh2021adversarial, thakoor2021bootstrapped,qiu2020gcc}. While these approaches effectively encode global information and generate valuable graph-level representations, a significant challenge lies in transferring knowledge from a specific pretext task to downstream tasks with substantial gaps, potentially resulting in negative transfer~\cite{rosenstein2005transfer}. This can limit the applicability and reliability of pre-trained models and potentially yield less favorable outcomes, even worse than learning from scratch.

\subsection{Multi-task Pre-training}\label{subsec:multi_pre}
Multi-task pre-training accommodates multiple optimization objectives, addressing a broad spectrum of graph-related aspects to enhance generalization while mitigating negative transfer issues. These objectives may encompass various combinations, such as concurrent training of node attribution reconstruction and structural recovery~\cite{hu2020strategies, hu2020gptgnn, zhang2020graphbert, 
fang2022geometryenhanced}. For example, Hu et al.~\cite{hu2020strategies} pre-trained a GNN at both the node and graph levels, enabling the GNN to learn valuable local and global representations simultaneously. Furthermore, some works have employed contrastive learning at different levels~\cite{jiang2021contrastive, xu2021infogcl}, or joint optimization of both contrastive loss and graph reconstruction error~\cite{li2023adaptergnn}. However, it is crucial to recognize that multi-task pre-training approaches may face optimization challenges, potentially resulting in suboptimal performance across tasks. As a result, optimizing model performance for each task while mitigating the negative transfer problem remains a significant but unresolved concern.

\subsection{Further Discussion} \label{subsec:pre-train-dis}
Fortunately, the prompting and predicting paradigm offers a robust solution to the challenges mentioned above. This approach can fully exploit model performance and seamlessly integrate with advanced GNN architectures. Instead of adapting pre-trained GNNs to downstream tasks through objective engineering, this paradigm reformulates downstream tasks into those solved during the pre-training phase using a graph prompt. This innovative strategy effectively bridges the gap between pretext and downstream tasks while sidestepping suboptimal performance pitfalls. Furthermore, in comparison to traditional fine-tuning approaches, the prompting paradigm showcases remarkable flexibility, enabling it to excel in scenarios demanding few-shot or even zero-shot learning, where adapting to new contexts with limited or no labeled data is paramount. In the current landscape marked by surging model volumes and an ever-expanding array of downstream tasks, the ascent of the prompting paradigm represents an irresistible and transformative trend.

\begin{table*}[ht]
\centering
\caption{Summary of existing representative works on graph prompt. $\mathcal{S}$: Subgraph. $V(\mathcal{S})$: Node set within subgraph $\mathcal{S}$. $\pi$: Pre-trained parameters. $\phi$: Task head parameters. $\theta$: Prompt parameters. $\tilde{\textbf{s}}$: Filled prompt.
}
\label{tab:graph_prompt_summary}
\resizebox{0.99\textwidth}{!}{
\begin{tabular}{@{}c|c|c|c|c|c|c|c|c|c|c|c@{}}
\toprule
                  & \multicolumn{3}{c|}{pre-training task}                                          
                  & \multicolumn{3}{c|}{prompt design}
                  & \multicolumn{3}{c|}{downstream tasks}   
                  & \multicolumn{2}{c}{answering function}   \\ \midrule
Paper             & \multicolumn{1}{c|}{node} & \multicolumn{1}{c|}{edge} & graph 
                  & prompt components & inserting pattern  & prompt tuning 
                  & \multicolumn{1}{c|}{node} & \multicolumn{1}{c|}{edge} & graph 
                   & \multicolumn{1}{c|}{Preset} & \multicolumn{1}{c}{Learnable}
                  \\ \midrule 
\makecell{GPPT \\(KDD 2022 \cite{sun2022gppt})} & \xmark &\cmark & \xmark & 
\makecell{structure token: \\$\textbf{s}_v \in \mathbb{R}^d$ \\task token: \\$\textbf{c}_y \in \mathbb{R}^d$}& 
\makecell{$\textbf{s}_{v_i} \leftarrow f_{\theta} (v_i)$  \\ $\tilde{\textbf{s}}_{y,v_i} \leftarrow [\textbf{c}_y, \textbf{s}_{v_i}]$}
&
Cross Entropy
&  \cmark & \xmark & \xmark & \cmark & \xmark \\\midrule

\makecell{GPF \\(arXiv \cite{fang2022prompt}) }& \cmark & \cmark & \cmark  
& prompt feature $\textbf{p} \in \mathbb{R}^d$ 
& $\tilde{\textbf{s}}_i \leftarrow \textbf{x}_i + \textbf{p}$  
& \makecell{$\max_{\textbf{p}, \phi} \sum_{(y_i,\tilde{\textbf{s}}_i)}$\\$ p_{\pi, \phi} (y_i| \tilde{\textbf{s}}_i) $} 
& \xmark   & \xmark  & \cmark & \xmark & \cmark \\ \midrule

\makecell{All in One\\ (KDD 2023 \cite{sun2023all})} & \xmark & \xmark & \cmark  
& \makecell{prompt token: \\$\mathcal{P}=\{ \textbf{p}_1, ..., \textbf{p}_{|\mathcal{P}|} \}$ \\token structure: \\$\{ (\textbf{p}_i, \textbf{p}_j) | \textbf{p}_i, \textbf{p}_j \in \mathcal{P} \}$}

&\makecell{$w_{ik} \leftarrow \sigma(\textbf{p}_k \cdot \textbf{x}^T_i)$ \\ if $\sigma(\textbf{p}_k \cdot \textbf{x}^T_i) > \delta$ else $0$  \\$\tilde{\textbf{s}}_i \!\!\!\leftarrow \textbf{x}_i \!\!+ \sum^{|\mathcal{P}|}_{k=1} w_{ik} \textbf{p}_k $ }

& 
Meta-Learning
&  \cmark   &\cmark & \cmark &\cmark &\cmark\\\midrule

\makecell{GraphPrompt \\(WWW 2023\cite{liu2023graphprompt})} & \xmark & \cmark & \xmark 
& \makecell{prompt token: \\$\textbf{p}_t \in \mathbb{R}^d, t \in \mathcal{T}$  \\structure token: $\textbf{s} \in \mathbb{R}^d$ \\task token: $\textbf{c}_y \in \mathbb{R}^d$}

& \makecell{
$\tilde{\textbf{s}}^t_i \leftarrow Readout(\{ \textbf{p}_t \odot f_{\pi} (v) | $\\$v \in V(\mathcal{S}_i) \})$ \\$\textbf{c}_y \leftarrow \textrm{Mean} (\{ \tilde{\textbf{s}}^t_j| y_j = y \}) $} 

&\makecell{$\min_{\textbf{p}_t} - \sum_{(y_i, \mathcal{S}_i)}\ln$\\$  \frac{\exp ( \textrm{sim} (\tilde{\textbf{s}}_i^t , \textbf{c}_{y_i} ) / \tau )}{ \sum_{y \in Y} \exp ( \textrm{sim} (\tilde{\textbf{s}}_i^t , \textbf{c}_y) / \tau) } $ }
&  \cmark   &\xmark  &\cmark & \cmark & \xmark \\  \midrule

\makecell{PGCL \\(arXiv \cite{gong2023prompt})} & \xmark & \xmark & \cmark 
& \makecell{semantic token: $\mathbf{p}^s\in \mathbb{R}^d$\\contextual token: $\mathbf{p}^c\in \mathbb{R}^d$}

& \makecell{$z_x^{p s}=z_x^s\odot \mathbf{p}^s$\\$ z_x^{p c}=z_x^c \odot \mathbf{p}^c$
} 

&\makecell{$\min-\sum_{(v, a, b) \in \mathcal{T}} \log $\\$\frac{\exp \left(\operatorname{sim}\left(z_v^p, z_a^p\right) / \tau\right)}{\sum_{u \in\{x, y\}} \exp \left(\operatorname{sim}\left(z_v^p, z_u^p\right) / \tau\right)}$ }

&  \cmark   &\cmark  &\cmark & \cmark & \xmark \\  \midrule

\makecell{PRODIGY \\(NeurIPS 2023 \cite{huang2023prodigy}) }& \xmark & \cmark & \xmark & 
\makecell{data graph: \\$\mathcal{G}^D \sim \oplus^k_{i=1} \mathcal{N} (\mathcal{V}_i,\mathcal{G)_i}$ \\task graph: $\mathcal{G}^T$}& 
$\tilde{\textbf{s}}_i \leftarrow f^T_{\pi^T} (\mathcal{G}^T | f^D_{\pi^{D}} (\mathcal{G}^D))$  & 
Fixed &  \cmark   & \cmark  & \cmark& \cmark & \xmark \\ \midrule

\makecell{SGL-PT \\(arXiv \cite{zhu2023sglpt}) }& \cmark & \xmark & \cmark & 
\makecell{prompt token: \\one vector for each graph}
& 
connect to all nodes in the graph  & 
\makecell{contrastive loss and \\reconstruction loss} &  \cmark   & \xmark  & \xmark& \cmark & \xmark \\ \midrule

\makecell{GPF-Plus \\(NeurIPS 2023 \cite{fang2023universal}) }& \cmark & \cmark & \cmark  

& \makecell{prompt features\\ $\textbf{p}_1,\cdots,\textbf{p}_k  \in \mathbb{R}^d$} 

& \makecell{$\tilde{\textbf{s}}_i \leftarrow \textbf{x}_i + \sigma(\textbf{p}_1,$\\$\cdots,\textbf{p}_k )$}  

& \makecell{$\max_{\textbf{p}, \phi} \sum_{(y_i,\tilde{\textbf{s}}_i)}$\\$ p_{\pi, \phi} (y_i| \tilde{\textbf{s}}_i) $ }

& \xmark   & \xmark  & \cmark & \xmark & \cmark \\ \midrule

\makecell{DeepGPT\\(arXiv \cite{shirkavand2023deep})} & \cmark & \cmark & \cmark 
& \makecell{prompt token: \\$\textbf{p} \in \mathbb{R}^d$ \\ prefix token: \\$\textbf{P} \in \mathbb{R}^{|\mathcal{P}| \times d}$ }

& \makecell{ $\tilde{\textbf{x}}_i \leftarrow \textbf{x}_i + \textbf{p} $ \\ $\tilde{\textbf{s}}_i \leftarrow  f_{\textbf{P}, \pi}(G,  \tilde{\textbf{x}}_i) $}

& \makecell{$\min_{\textbf{p}, \textbf{P}, \phi} \sum_{(y_i, v_i)}$\\$ \mathcal{L}( p_{\phi}(\tilde{\textbf{s}}_i), y_i )  $}

& \xmark  & \xmark & \cmark & \xmark & \cmark \\ \midrule

\makecell{ULTRA-DP \\(arXiv \cite{chen2023ultradp})}& \xmark & \cmark &\xmark  
& \makecell{prompt token: \\$\textbf{p}_{i}=\textbf{p}^{\textrm{task}} + w^{\textrm{pos}}\textbf{p}_i^{\textrm{pos}}$, \\$\textbf{p}_i^{\textrm{pos}}$ denotes $v_i$'s \\positional embedding}

& \makecell{create a virtual node \\$v_i^p$ for target node $v_i$,\\ 
$G' \leftarrow ( \mathcal{V}\cup \{ v_i^p\},$  \\
$ \mathcal{E}\cup \{ (v_i^p, v_i) \},\textbf{X} \cup \{ \textbf{p}_i \})$
}

& \makecell{Multi-task-Learning} 

&\cmark   & \xmark  & \xmark& \xmark & \cmark \\ \midrule

\makecell{HetGPT \\(arXiv \cite{ma2023hetgpt})} & \xmark& \cmark & \xmark  
& \makecell{prompt token: \\$\mathcal{F}= \{ \textbf{f}_i^A \}_{i=1}^K$ \\for $A \in \mathcal{A}$\\ task token: $\textbf{c}_y \in \mathbb{R}^d$}

&  \makecell{$\tilde{\textbf{s}}_i^A \leftarrow \textbf{x}_i^A +$\\$ \sum_{k=1}^K w_{ik} \textbf{f}_k^A$, \\ $\textbf{z}_i = f_{\pi}(G, \tilde{\textbf{s}}_i^A)$}

& \makecell{$  \min_{\textbf{C}, \mathcal{F}} -\sum_{(y_i, v_i)} \log $\\$  \frac{  \textrm{exp}( \textrm{sim} ( \textbf{z}_i, \textbf{c}_{y_i} )/\tau )  }{ \sum_{y \in \textbf{Y}}  \textrm{exp}( \textrm{sim} ( \textbf{z}_i, \textbf{c}_{y} ) /\tau )  }       $ }

&\cmark   & \xmark  &\xmark & \cmark & \xmark \\ \midrule

\makecell{SAP\\(arXiv \cite{ge2023enhancing})} & \cmark & \xmark & \xmark 
& \makecell{task token: \\$\mathcal{P}= \{ \textbf{c}_y \}_{y \in \textbf{Y}}$ \\structure token: \\$\mathcal{W}= \{ (v_i, c_j) \}_{v_i \in \mathcal{V}, c_j \in \mathcal{P}}$ }

& \makecell{$G' \leftarrow (\mathcal{V} \cup \mathcal{P}, \mathcal{E} \cup \mathcal{W})$ \\$\textbf{Z}^{(1)} = \textrm{MLP}_{\pi'}(\textbf{X})$ \\ $\textbf{Z}^{(2)} = \textrm{GNN}_{\pi''}( [ \textbf{X}, \textbf{P}], $\\$[\textbf{A}, \textbf{W}] )$}

& \makecell{$\min_{\textbf{W}}  - \sum_{(y_i,v_i)} \log $\\$\frac{\textrm{exp}( \textrm{sim}( \textbf{z}_i^{(1)}, \textbf{z}_{y_i}^{(2)} ) / \tau )}{ \sum_{y \in \textbf{Y}} \textrm{exp}( \textrm{sim}( \textbf{z}_i^{(1)}, \textbf{z}_{y}^{(2)} ) / \tau ) }$}

& \cmark  &\xmark& \cmark& \cmark & \xmark \\ \midrule

\makecell{VNT\\(KDD 2023 \cite{tan2023virtual})} & \cmark & \cmark & \xmark 
& \makecell{$\boldsymbol{P}=\left[\boldsymbol{p}_1 ; \ldots ; \boldsymbol{p}_P\right]$,\\$\boldsymbol{p}_p \in \mathbb{R}^F$ }

& \makecell{$\left[E^1 \| Z^1\right]=L^1\left(\left[E^0 \| P\right]\right) $\\$\in \mathbb{R}^{(V+P) \times F}$}

& \makecell{Cross Entropy} 

& \cmark &\xmark& \xmark& \xmark & \cmark \\

\bottomrule
\end{tabular}%
}
\end{table*}

\section{Prompt Design for Graph Tasks}\label{sec:ptask}
\begin{figure}[t]
    \centering
    \includegraphics[width=0.49\textwidth]{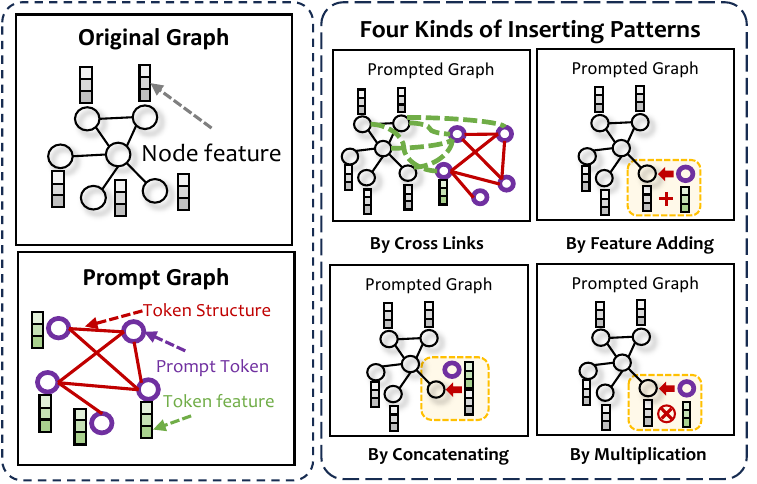}
    \caption{Prompt Tokens, Structures, and Inserting Patterns.}
    \label{fig:prompt}
\end{figure}

In this section, we propose a unified view for the graph prompt. As shown in \figref{fig:prompt}, the graph prompt should contain at least three components: prompt tokens with prompt vector; token structures preserving inner correlations of these tokens; and inserting patterns indicating how to integrate the original graph with prompts. Beyond these details, we are particularly interested in the following questions: \textit{Question 1:} How do these works design the graph prompt? \textit{Question 2:} How do these works reformulate downstream tasks to the pre-training tasks? \textit{Question 3:} How do these works learn an effective prompt? and \textit{Question 4:} What are the inner connections of these works, their advantages and limitations? With these questions, we summarize the most representative works published recently and present them in \tabref{tab:graph_prompt_summary}.

\subsection{Prompt Token, Structure, and Inserting Pattern}\label{subsec:pdesign}

\textit{\dotuline{Question 1: How do They Design A Graph Prompt?}}

\textbf{A. Prompt as Tokens. }
The simplest graph prompt can be treated as some additional features added to the original graph features \cite{zhu2023graphcontrol}. For example, given a graph feature matrix $\mathbf{X}=\{x_1,\cdots, x_N\} \in \mathbb{R}^{N\times d}$ where $x_i\in \mathbb{R}^{1\times d}$ is the feature of $i$-th node and $d$ is the dimension of the feature space. \citet{fang2022prompt} and \citet{shirkavand2023deep} treat the basic prompt as a learnable vector $p \in \mathbb{R}^{1\times d}$, which can be added to all node features and make the manipulated feature matrix be $\mathbf{X}^{*}=\{x_1+p,\cdots, x_N+p\}$. In this way, we can use the reformulated features to replace the original features and process the graph with pre-trained graph models. A later work \cite{fang2023universal} further extends one prompt token to multiple tokens and makes the performance better. PGCL \cite{gong2023prompt} design a prompt vector for semantic view and another prompt vector for contextual view, then they add these prompt vectors to the graph-level representations by element-wise multiplication. A similar prompt design is also adopted in VNT \cite{tan2023virtual}. The difference is that their inserting pattern does not add the prompt token to the original graph feature but concatenates the prompt token with the original node set and tries to integrate them by the self-attention function in a graph transformer. In GraphPrompt \cite{liu2023graphprompt}, the prompt token is similar to the format defined by \citet{fang2022prompt}. The difference thing is that previous work designed the prompt tokens in the initial feature space, while this method assumes the prompt in the hidden layer of the graph model. Usually, the hidden size will be smaller than the original feature, making their prompt token shorter than the previous one. Another different thing is that the graph prompt here is used to assist graph pooling operation (a.k.a $\text{Readout}$). For example, given the node set $\mathcal{V}=\{v_1,\cdots, v_{|\mathcal{V}|}\}$, the embedding of node $v$ is $\textbf{h}_v \in \mathbb{R}^{1\times d}$, a prompt token $\textbf{p}_t \in \mathbb{R}^{1\times d}$ specified to task $t$ is inserted to the graph nodes by the element-wise multiplication ($\otimes$): $\textbf{s}_t=\text{Readout}(\{\textbf{p}_t \otimes \textbf{h}_v:v \in \mathcal{V}\})$. Similarly, SGL-PT \cite{zhu2023sglpt} creates a prompt token to connect to all nodes in the graph. The prompt token preserves a global perception of the graph and can assist their global branch of the pre-training tasks, which can be also treated as a pooling strategy. Aiming at aligning node classification and link prediction, GPPT \cite{sun2022gppt} defines graph prompts as additional tokens that contain task tokens and structure tokens. Here the task token refers to the description of the downstream node label to be classified and the structure token is the representation of the subgraph surrounding the target node. By this means, predicting node $v$'s label can be reformulated to predict a potential link between node $v$'s structure token and the label task token. 
Aiming at the node classification task, ULTRA-DP \cite{chen2023ultradp} creates a prompt token for each target node, where the token feature is the weighted sum of position embedding of the target node and a task embedding of the pre-training task. HetGPT \cite{ma2023hetgpt} design a prompt with node tokens and class tokens, which are organized in a similar way to GPPT. The difference is that they also add a type-specific feature token to make graph prompts sensitive to different node types, by which they can extend existing graph prompts to heterogeneous graphs.

\textbf{B. Prompt as Graphs. }
The graph prompt in All in One \cite{sun2023all} is an additional subgraph that can be learned by efficient tuning. The prompt tokens are some additional nodes that have the same size of node representation as the original nodes. They assume the prompt tokens should be in the same semantic space as the original node features so that we can easily manipulate node features with these tokens. The token structures include two parts. The first part is the inner links among different tokens, and the second part is the cross-links between the prompt graph and the original graph. These links can be pre-calculated by the dot product between one token to another token (inner links) or one token to another original node (cross links). The inserting pattern is to add the prompt graph to the original graph by the cross-links and then treat the combined graph as a new graph, and send it to the pre-trained graph model to get the graph-level representation. The prompt graph in PRODIGY \cite{huang2023prodigy} includes data graphs and a task graph. The data graphs can be treated as subgraphs surrounding the target nodes (for node classification task), node pairs (for edge classification task), or just denoted as the graph classification instance. Here the prompt tokens and prompt structures are just the same as in the original graph. The task graph contains data tokens and label tokens where each data token connects to one data graph and is further connected by label tokens. Unlike previous works that aim at reformulating downstream tasks to the pre-training task by prompting the downstream data, PRODIGY leverages the prompt graph to unify all the upstream and downstream tasks. Their pre-training strategy is a set of tasks including neighboring matching and label matching, which can be reformulated as predicting the similarity between the data token and the label token in the prompt graph. SAP \cite{ge2023enhancing} also connects several prompt tokens (each token corresponds to one class) to the original graph by cross-links defined in All in One. The difference is that their prompting task is a node-level contrastive task, in which they use MLP to encode the node features as the first view and they use a GNN to encode the prompted graph as the second view, which is consistent with their pre-training task.

\subsection{Aligning Tasks by Answering Function}\label{subsec:align}

\textit{\dotuline{Question 2: How do they reformulate downstream tasks to the pretext?} }

\textbf{A. Handling Different Level Tasks.} 
All in One \cite{sun2023all} pre-trains the graph model via graph-level contrastive learning. The pre-training task aims to learn a robust graph encoder over different graph views generated by the perturbations of the original graph. To reformulate various graph tasks to this graph-level pretext, they first unify node-level, edge-level, and graph-level tasks to the graph-level task by induced subgraphs, which are also introduced in PRODIGY \cite{huang2023prodigy,liu2023graphprompt,gong2023prompt}. Then they claim that the graph prompt added to the original graph is in nature the simulation of any graph operations such as node feature masking, node or edge perturbations, subgraph removing, etc. With this opinion, we just need to add an appropriate graph prompt to downstream graph datasets then it will be promising to further reformulate the downstream task to the pretext.

\textbf{B. Learnable Answering Function.} 
To output the results of downstream tasks, \citet{sun2023all} design two types of answering functions. The first one is a learnable task head (such as an MLP mapping function) that can be easily tuned with very limited data. It takes the graph-level representation generated by the pre-trained graph model and then outputs the downstream result. For example, if the downstream is a three-class node classification, we can simply use a dense layer with three hidden units to connect the graph representation, which is generated by the pre-trained model on the combined graph with node included graph and a graph prompt. In this case, both the graph prompt and the task head are tunable, so we can adjust them alternately. Similar learnable answering functions are also adopted in other works like \cite{fang2022prompt, fang2023universal,tan2023virtual}. The good point is that they are very easy to align two tasks, however, it also increases the tuning workflow.

\textbf{C. Hand-crafted Answering Function.} 
To further reduce the tuning burden, All in One \cite{sun2023all} also proposes a second answering function, which is hand-crafted without any trainable task head. For example, for a node classification task, we can set up $K$ unique sub-prompts, each aligning with a different node type, where $K$ represents the total number of node categories. If the pre-training involves a task like GraphCL \cite{you2020graph}, which aims to maximize similarity at the graph level between pairs of graph views, then the target node can be classified with label $\ell, (\ell=1,2,\cdots,K)$ if the $\ell$-th graph most closely resembles the original node-inclusive graph. Similarly, GPPT \cite{sun2022gppt} and  HetGPT \cite{ma2023hetgpt} use link prediction as their pre-training task and reformulate downstream node classification by unifying it as the same task template. For example, by treating the node label as an additional token, we can use the pre-trained model to directly output the possibility of an edge between the label token and the target node. 
The pre-training strategy in SAP \cite{ge2023enhancing} is to compare node representations from two graph views, the first of which is generated by node feature encoding and the second of which is encoded by a graph model. To this end, their prompt tokens denote class information and they compare node representation with each class token to find the class with the largest similarity as the inference results. By designing a unified task template, PRODIGY \cite{huang2023prodigy} uses a hand-crafted graph prompt to describe all node, edge, and graph classification tasks by predicting the link between data tokens and label tokens. \citet{liu2023graphprompt} extend the link prediction task as graph pair similarity and treat it as their pre-training task, to align the downstream node classification and graph classification task, they design a unified answering template making the downstream side aligned with the pre-training side. Specifically, given a triplet of induced graphs $(g_1,g_2, g_3)$ where $g_1$ and $g_2$ have the same label, $g_1$ and $g_3$ have different labels. In particular, when the target task is node classification, the induced graph refers to the contextual subgraph of the target node. The unified answering template is defined as  $\text{sim}(g_1,g_2)>\text{sim}(g_1,g_3)$.

\subsection{Prompt Tuning}\label{subsec:pt}

\textit{\dotuline{Question 3: How do They Learn A Graph Prompt? }}

\textbf{A. Meta-Learning Technique. }
To learn appropriate prompts, \citet{sun2023all}  leverage meta-learning techniques (such as MAML\cite{finn2017modelagnostic} model) to obtain a robust starting point for the prompt parameters. Since the support set and query set include various graph tasks (such as node classification, link prediction, graph classification, etc), the learned graph prompt is expected to be more general on various downstream tasks. 

\textbf{B. Task-specific Tuning. }
Besides All in One \cite{sun2023all}, which aims to learn a general prompt on various downstream tasks, there are also some works that target specific downstream tasks such as graph classification. In this case, the prompt tuning can be more task-directed. For example, GPF \cite{fang2022prompt} aims at a graph classification task, so it just needs to tune the prompt token $\textbf{p}$ and the task head $\phi$ by maximizing the likelihood of predicted correct graph labels given the prompted graph representation $\tilde{\textbf{s}}_i$ from the pre-trained graph model $\pi$. In this case, the task head tuning and the prompt tuning share the same objectives, which can be formulated by: $\max_{\textbf{p}, \phi} \sum_{(y_i,\tilde{\textbf{s}}_i)} p_{\pi, \phi} (y_i| \tilde{\textbf{s}}_i)$.

\textbf{C. Tuning in Line with Pretext. }
Intuitively, prompting aims at reformulating downstream tasks to the pre-training task. Therefore, it would be more natural if the prompt tuning shares the same objective with the pre-training task. As suggested in GraphPrompt \cite{liu2023graphprompt}, the authors use a similar loss function to learn prompts. Similarly, GPPT \cite{sun2022gppt} and VNT \cite{tan2023virtual} adopt the same loss function (Cross-Entropy) as their link prediction and node classification tasks, respectively. HetGPT \cite{ma2023hetgpt} and  SAP \cite{ge2023enhancing} use a node-level contrastive loss to learn their prompt tokens because their pre-training task is also conducted by the same contrastive task (node pair comparison). PGCL \cite{gong2023prompt} introduces graph-level loss to align with the pre-training task. ULTRA-DP \cite{chen2023ultradp} develop two pre-training tasks including edge prediction and neighboring prediction, each of which corresponds to one task embedding. In the pre-training phase, they randomly select a task and then integrate specific task-related embeddings into the prompt tokens. These learnable task embeddings are then trained with the graph model.

\subsection{Further Discussion}\label{subsec:pdiscussion}

\textit{\dotuline{Question 4: What are Their Connections, Pros and Cons?}}

The good point of GPF \cite{fang2022prompt} is that they propose a very simple prompt that can be easily used in various pre-training tasks and downstream tasks. However, a single prompt token added to all nodes is very limited in expressiveness and generalization. A potential solution is to learn an independent prompt token for each node, which means one node corresponds to one prompt token, but this will cause low efficiency in parameters. To this end, we can train K-independent basis vectors and use them to compound each node token (GPF-Plus \cite{fang2023universal}). This improvement makes their work have more similar insights with All in One \cite{sun2023all}.

HetGPT \cite{ma2023hetgpt} extends prompt tokens to type-specific format, which can deal with graph prompting in heterogeneous data. However, they can only deal with node classification tasks. To this end, GraphPrompt \cite{liu2023graphprompt} reformulates link prediction to graph pair similarity task. It is worth noticing that the role of their prompt token is very similar to the project vector in the graph attention network. There are also some attention-based graph-pooling methods, which share the same motivation. The difference is that the authors claim the graph-pooling component in the pre-training stage might not fit other downstream tasks, thus needing additional prompts to redirect the graph-pooling component in the graph model.  

GPPT \cite{sun2022gppt} represents a specific instance within the broader framework of All in One \cite{sun2023all}. For instance, if we minimize the prompt graph to isolated tokens that correlate with node classes and substitute the resulting graphs with a complete graph, the All in One prompt structure can be simplified into the GPPT format. This allows for the utilization of edge-level pretexts in node classification tasks within the GPPT framework. The shortcoming of GPPT might be that it is restricted to binary edge prediction pretexts and is solely effective for node classification in downstream tasks. In comparison, frameworks like GraphPrompt and All in One are designed to accommodate a wider array of graph-related tasks, extending beyond just node-level classification. The good point is that when adapting models for different tasks, GraphPrompt, GPF, and GPF-Plus often require the tuning of an extra task-specific module. In contrast, All in One, and GPPT utilize task templates that focus more on the manipulation of input data and are less dependent on the specifics of downstream tasks.

Intuitively, the data graphs, one of the components in PRODIGY \cite{huang2023prodigy}, are very similar to the induced graph in All in One and GraphPrompt. The pre-training task in PRODIGY can be seen as predicting a link between the data token and the label token, which shares a similar insight with GPPT. The good thing is that their prompts have no trainable parameters, which means they do not need to tune the prompt and are more efficient in the in-context learning area. PRODIGY  does not need any tuning work and can be used in knowledge transferring between different datasets. However, a non-tunable prompt is usually not flexible enough, which might also limit the generalization of the pre-trained model when the downstream tasks to be transferred are located in a different domain from the pre-training one. In contrast, ULTRA-DP \cite{chen2023ultradp} tune prompt both in the pre-training stage and the downstream tasks. It first put the prompt tuning work in the pre-training stage to obtain the task embeddings, which are one of the main components in their prompt. Then they use these task embeddings to initialize a downstream prompt. Intuitively, their prompts are not used to reformulate downstream tasks to the pretext. Instead, these prompt tokens are used to select suitable pre-training tasks from a task pool to fit the downstream task. It is still an interesting question of how to achieve the optimal balance given efficiency, generalization, and the flexibility of prompt.

Compared with other works that usually define clear inserting patterns, VNT \cite{tan2023virtual} concatenates prompt tokens with the original node set and then puts all of them into the graph transformer. Actually, the graph transformer will leverage a self-attention function to further calculate the correlations among them, which can also be treated as a variant of inserting patterns defined in All in One. The good thing is that we do not need to design a threshold to tailor the connection but the shortcoming is that it can only use a graph transformer as its backbone and can not applied to more message-passing-based graph models. In addition, there are also some more advanced variants of graph transformers requiring additional position embedding as one of their input. However, the prompt tokens in VNT have no clear inserting links to the original graph, which might not make it easy to apply existing position encoding approaches for these graph transformer variants.

\section{Graph Prompting in Multi-Modal and Multi-Domain Areas}\label{subsec:pmodal}

\subsection{Multi-Modal Prompting with Graphs}\label{subsec:multi_modal}

The fusion of images, sound, and text has been widely studied and achieved remarkable success. However, most of these modalities are described by linear data structure. In our real-world life, there are more kinds of data in non-linear structures like graphs. How to connect these linear modalities (e.g. text, images, sound, etc) to the non-linear modalities (e.g. graphs) has become a very attractive research topic because it is a bigger move towards artificial general intelligence. Unfortunately, reaching this vision is very tough. Currently, we only see some hard progress in the fusion of text and graphs, especially in the text-attributed graphs. With the help of recent large language models, the fusion of text and graph has achieved even more notable performance. Since there have already been some informative surveys on this topic, we next briefly discuss some representative works that are closely related to \textit{\dotuline{prompt techniques}}. We suggest readers refer to the mentioned literature \cite{li2023survey,liu2023graph} to require more detailed information further.

\textbf{A. Prompt in Text-Attributed Graphs.} \citet{wen2023augmenting} ventured into enhancing text classification in scenarios with limited data resources by the proposed model, Graph-Grounded Pre-training and Prompting (G2P2). Their work identifies the issue of insufficient labeled data in supervised learning and proposes a solution leveraging the inherent network structure of text data, such as hyperlinks or citation networks. G2P2 utilizes text-graph interaction strategies with contrastive learning during the pre-training phase, followed by an inventive prompting technique during the classification phase, demonstrating notable effectiveness in zero- and few-shot text classification tasks. \citet{zhao2023gimlet} focused on molecule property prediction, a field grappling with the scarcity of labeled data. Their study introduces an integrated graph-text model to enhance prompt-based molecule task learning in a zero-shot context. This model employs generalized position embedding and decouples encoding of the graph from task prompt, enhancing its generalization capability across novel tasks. \citet{li2023promptbased} explored node classification within the framework of multi-modal data (text and graph), particularly focusing on limited-label scenarios. Unlike traditional works that usually feed pre-computed text features into graph neural networks, they incorporate raw texts and graph topology by a hand-crafted language prompt template into the model design.

\textbf{B. Large Language Models in Graph Data Processing. }
\citet{chen2023exploring} explored the potential of Large Language Models (LLMs) in graph node classification tasks. They investigated two pipelines: LLMs-as-Enhancers, which enhances node text attributes using LLMs followed by predictions via Graph Neural Networks (GNNs), and LLMs-as-Predictors, which directly employs LLMs as standalone predictors. Their empirical evaluations revealed that deep sentence embedding models and text-level augmentation through LLMs effectively enhance node attributes, while LLMs also show promise as standalone predictors, albeit with concerns about accuracy and test data leakage. \citet{fatemi2023talk} conducted a comprehensive study on encoding graph-structured data for consumption by LLMs. Their findings include the influence of the graph encoding method, the nature of the graph task, and the structure of the graph on LLM performance. They demonstrated that simple prompts are most effective for basic graph tasks and that graph encoding functions significantly impact LLM reasoning. Their experimental setup introduced modifications to the graph encoding function, revealing improvements in performance and demonstrating the effect of model capacity on graph reasoning ability. \citet{jin2023patton} introduced an innovative approach to pre-train language models on networks rich in text. Their framework, named PATTON, focuses on integrating the intricacies of textual attributes with the underlying network structure. They developed two novel pretraining strategies: one concentrating on the context within networks for masked language modeling, and the other on predicting masked nodes, thus capturing the interplay between text and network structure. The effectiveness of PATTON was validated through various experiments, showcasing its superior performance over traditional text/graph pretraining methods in diverse tasks such as document classification, retrieval, and link prediction in different domain datasets. This approach signifies a shift in pretraining methodologies, emphasizing the synergy between textual data and network context. \citet{wang2023knowledge} proposed a Knowledge Graph Prompting (KGP) method to enhance LLMs for multi-document question answering (MD-QA). They created a knowledge graph over multiple documents, with nodes representing passages or document structures and edges denoting semantic/lexical similarity. The LM-guided graph traverser in KGP navigates the graph to gather supporting passages, aiding LLMs in MD-QA. Their experiments indicated that the construction of KGs and the design of the LM-guided graph traverser significantly impact MD-QA performance.

\textbf{C. Multi-modal Fusion with Graph and Prompting. }
The integration of multi-modal data using graph and prompting techniques has seen remarkable progress in recent years. For example, \citet{edwards2023synergpt} propose SynerGPT in the field of drug synergy prediction. This model leverages a transformer-based approach, uniquely combining in-context learning with genetic algorithms to predict drug synergies. In the area of vision-language models, \citet{li2023graphadapter} develop GraphAdapter, a prompt-based strategy that utilizes an adapter-style tuning mechanism, bringing together textual and visual modalities through a dual knowledge graph. \citet{liu2023gitmol} extend work multi-modal fusion into molecular science with their proposed GIT-Mol. A large language model integrates graph, image, and textual data with the help of prompt, offering substantial improvements in various tasks like molecule generation and property prediction. Although much effort has been made in the past few years, the academic is still trying hard to find better solutions to integrate text and graphs via text-attributed graphs or knowledge graphs. There is still a very large imagination in the fusion of more kinds of modalities.

\subsection{Graph Domain Adaptation with Prompting}\label{subsec:pdomain}
The field of graph domain adaptation has seen significant advancements, particularly with the integration of prompting techniques. However, graph domain adaptation is still not a well-solved problem because there exist at least two challenges: The first one is how to align semantic spaces from different domains. The second one is how to identify structural differences.

\textbf{A. Semantic Alignment. }
In particular, All in One \cite{sun2023all} extends the ``pre-training and fine-tuning'' workflow with multi-task prompting for GNNs, unifying prompt formats, and introducing meta-learning for prompt optimization. To make the graph model adaptive to different graph domains, they first reveal that the graph prompt in nature can be seen as graph operation and then they use graph prompt to manipulate different domain graph datasets. GraphControl \cite{zhu2023graphcontrol} introduces a unique deployment module inspired by ControlNet, effectively integrating downstream-specific information as conditional inputs to enhance the adaptability of pre-trained models to target data. This approach aligns input space across various graphs and incorporates unique characteristics of the target data. \citet{zhang2023structure} presents a pre-training model for knowledge graph transfer learning. This model uses a general prompt-tuning mechanism, treating task data as a triple prompt, enabling flexible interactions between task KGs and task data. \citet{yi2023contrastive} combines personalized graph prompts with contrastive learning for efficient and effective cross-domain recommendation, particularly in cold-start scenarios. A representative work is proposed by \citet{liu2023one}, in which they describe graph nodes from different domains by language and then use LLM to get a textual embedding. However, this work needs the semantic name of each feature while sometimes graph features are usually latent vectors without clear semantic meaning.

\textbf{B. Structural Alignment.}
 \citet{cao2023when} analyze the feasibility between different graph datasets and found that the structural gap holds the upper bound of various graph model transferability. GraphGLOW \cite{zhao2023graphglow} addresses the limitations of existing models that operate under a closed-world assumption, where the testing graph is identical to the training graph. This approach often leads to prohibitive computational costs and overfitting risks due to the need for training a structure learning model from scratch for each graph dataset. To this end, it coordinates a single graph-shared structure learner with multiple graph-specific GNNs to capture generalizable patterns of optimal message-passing topology across datasets. The structure learner, once trained, can produce adaptive structures for unseen target graphs without fine-tuning, thereby significantly reducing training time and computational resources. AAGOD by \citet{guo2023datacentric} proposes a data-centric framework for OOD detection in GNNs. They use a parameterized amplifier matrix, which is treated as a prompt, to superimpose on the adjacency matrix of input graphs. Inspired by prompt tuning, \citet{shirkavand2023deep} propose DeepGPT for graph transformer models. They add prompt tokens to the input graph and each transformer layer. By updating the prompt tokens, they can efficiently tune a graph model on different graph datasets. 

\section{Potential Applications}\label{sec:app}
With the widespread utilization of networks as a data modeling structure for representing diverse relational information across social, natural, and academic domains, the graph prompt mechanism exhibits substantial potential for a wide range of real-world applications. In this section, we explore the potential applications of graph prompting in online social networks, recommender systems, knowledge management, and biology.

\textbf{Online Social Networks. }
Online social platforms consist of users who can be represented as nodes, and their social connections form online social networks (OSNs). Previous research has investigated the potential of prompting mechanisms in identifying fake news within OSNs to prevent malicious attacks \cite{wu2023promptandalign}. Specifically, they employ textual prompts applied to pre-trained language models (PLMs) to distill general semantic information. By combining this semantic signal with the dynamics of information propagation within social networks, improved classification performance can be achieved. While the use of tailored textual prompts for PLMs has been studied, the application of graph prompting mechanisms within social networks is still under-explored. In the future, it is promising to directly apply prompt tuning techniques to social networks, utilizing few-shot labels for tasks such as fake news detection or anomaly detection \cite{wen2023voucher, guo2023datacentric}, where the labeling process is laborious and requires domain expertise. By incorporating prompts directly within social networks, this approach can address the scarcity of labeled data and enhance the security and trustworthiness of online social networks.

\textbf{Recommender Systems. }
E-commerce platforms provide a valuable opportunity to leverage recommender systems for enhancing online services. While prompt tuning in recommender systems has received limited research attention, it holds significant potential \cite{yi2023contrastive,yang2023empirical, wu2023personalized, hao2024motifbased}. In \cite{yi2023contrastive}, the graph prompt tuning technique is applied to cross-domain recommendation scenarios to address the challenges of domain adaptation. Specifically, when applying a pre-trained recommendation model to the target domain, extra prompt nodes are introduced to achieve both efficient and effective domain recommendation. Meanwhile, \citet{yang2023empirical} propose personalized user prompts to bridge the gap between contrastive pretext \cite{wu2021selfsupervised, yu2022are} to downstream recommendation task. They design different kinds of personalized prompts, in combination with pre-trained user embeddings to facilitate dynamic user representations, leading to more accurate and personalized recommending results. In the future, further exploration into the integration of graph prompt tuning within recommender systems can be conducted to enhance recommendation performance, personalization, and adaptability across different domains.

\textbf{Knowledge Management. }There are two branches of research focused on performing prompting on knowledge graph (KG) for improved knowledge management. The first branch involves direct prompting on knowledge graphs using a pre-trained KG model to facilitate knowledge transfer, enabling better generalization across different KG data and tasks \cite{zhang2023structure}. The second branch explores the combination of grounded knowledge from KGs with the cognitive abilities of LLMs \cite{wang2023knowledge, tian2023graph, robinson2023leveraging, park2023graphguided, zhang2023benchmarking, pan2023unifying} to improve performance in downstream tasks. In the first research line, \citet{zhang2023structure} proposed a structure pre-training and prompt tuning approach to realize knowledge transfer. They designed specific pre-training objectives to obtain a powerful KG model. Subsequently, a general prompt tuning technique was employed to facilitate knowledge transfer between task-specific KGs and data. In the second research line, prompt tuning techniques are adopted to combine the grounded knowledge inherent in KGs with LLMs for enhancing downstream tasks. For example, in \cite{tian2023graph}, a novel graph neural prompting method was introduced to adapt KGs for LLMs by distilling valuable knowledge from KGs in a time- and parameter-efficient manner. Future research can further explore this trend by implementing more efficient graph prompting methods to fully distill beneficial knowledge from KGs to assist LLMs on diverse downstream tasks.

\textbf{Biology. }Molecules can be represented naturally as graphs, where atoms serve as nodes and chemical bonds act as edges \cite{rong2020selfsupervised}. Such graph modeling provides a basis for applying graph representation learning methods to perform tasks such as molecular property prediction, thereby benefiting scientific research and discovery \cite{rong2020selfsupervised, liu2023molca, zhao2023gimlet, edwards2023synergpt}. Previous research on graph representation learning in the molecular domain followed a task-specific approach. It involved training individual models tailored to specific molecule datasets \cite{rong2020selfsupervised, sun2020infograph}, lacking the generalization ability within the domain. Though there exist works that explored the utilization of LLMs (or LMs) as a universal tool for understanding molecules \cite{qian2023can, liu2023one}, these approaches primarily rely on regular texts to describe molecules, overlooking the inner graph structures within molecules \cite{zhang2023large}. Meanwhile, some recent works have focused on investigating the co-modeling of graphs and languages to preserve both structural dependencies and achieve generalization abilities across tasks and datasets, even under few-shot or zero-shot settings \cite{liu2023gitmol, zhao2023gimlet, edwards2023synergpt}. For instance, in \cite{liu2023gitmol}, the authors employed LoRA \cite{hu2021lora} to efficiently adapt to downstream tasks. Following this research direction, we believe that graph prompting techniques can also be adopted to achieve more efficient task adaptation within the molecular domain.

\section{\texttt{ProG}: A Unified Library for Graph Prompting}\label{sec:prog}
An indispensable component for fortifying the graph prompting ecosystem is a well-crafted tool. Despite the plethora of tools proposed for generalized graph learning, a notable absence persists in the realm of libraries dedicated to graph prompt functionalities. Addressing this gap, we are introducing \texttt{ProG} (Prompt Graph), an open-source, unified library meticulously designed to cater to the specific needs of graph prompting. This initiative promises to significantly enhance the landscape of graph-based applications by providing a versatile and comprehensive resource for researchers and practitioners alike.

\begin{figure}[!t]
    \centering
    \includegraphics[width=0.5\textwidth]{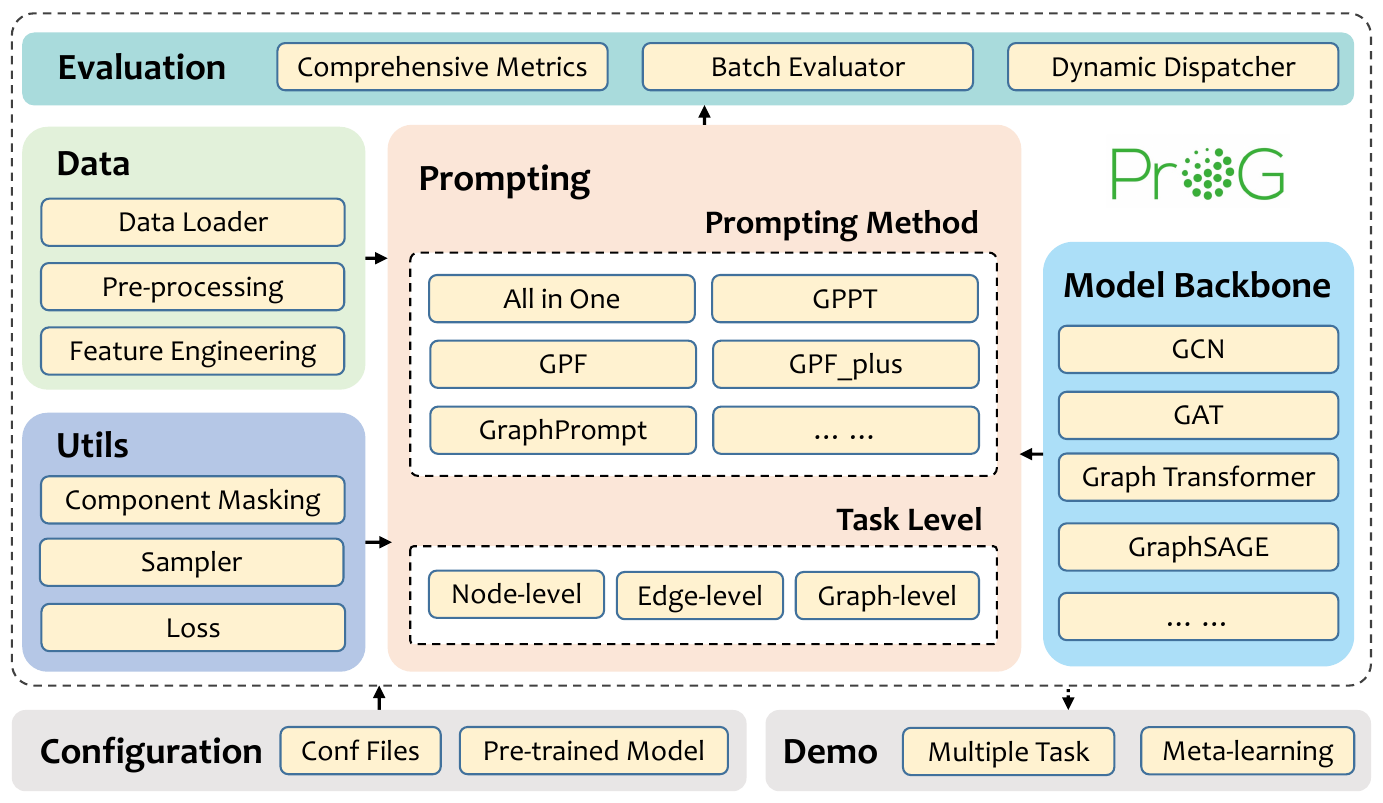}
    \caption{The architecture of \texttt{ProG}.}
    \label{fig:prog}
\end{figure}

\texttt{ProG} is a PyTorch-based library designed to facilitate single or multi-task prompting for pre-trained GNNs. 
The architecture is illustrated in Figure~\ref{fig:prog}.
It seamlessly integrates several widely used datasets in the graph prompt evaluation, including Cora, CiteSeer, Reddit,
Amazon,
and Pubmed etc.
The tool is equipped with essential evaluation metrics such as Accuracy, F1 Score, and AUC score, commonly employed in various graph prompt-related tasks. Notably, \texttt{ProG} incorporates state-of-the-art methods like All in One~\cite{sun2023all}, GPPT~\cite{sun2022gppt}, GPF~\cite{fang2022prompt}, and GPF-Plus ~\cite{fang2023universal}, and it continues integrating more graph prompt models. In summary, \texttt{ProG} offers the following key features:

\begin{itemize}
    \item \textbf{Quick Initiation.} All models are implemented within a consistent environment, accompanied by detailed demos, ensuring a swift initiation for newcomers.
    \item \textbf{Fully Modular.} \texttt{ProG} adopts a modular structure, empowering users to customize and construct models as needed.
    \item \textbf{Easy Extendable.} \texttt{ProG} is designed for seamless extension to encompass additional methods and a broader spectrum of downstream tasks, adapting to evolving research needs.
    \item \textbf{Standardized Evaluation.} \texttt{ProG} establishes a uniform set of evaluation processes, promoting equitable performance comparisons across models.
\end{itemize}

For additional information and access to the library, please visit the website of our library\footnote{\url{https://github.com/sheldonresearch/ProG}}. Additionally, we have curated a GitHub repository\footnote{\url{https://github.com/WxxShirley/Awesome-Graph-Prompt}}, serving as a centralized resource for the latest advancements in graph prompt learning. This repository includes a list of research papers, benchmark datasets, and available codes, fostering an environment conducive to ongoing research in this dynamic field. Regular real-time updates ensure that the repository remains current with emerging papers and associated codes.
\section{Challenges and Future Directions}\label{sec:dis}
\subsection{Current Challenges}\label{sec:challenges}
Graph prompt learning has made significant research progress, but it still encounters several challenges. In this subsection, we will discuss current challenges in detail.

\textbf{Inherent Limitation within Graph Models} Prompts, originated from the NLP field \cite{brown2020language, lester2021power, liu2023pretrain}, serve as a means to unlock the potential of pre-trained language models for adapting to downstream tasks. This in-context learning ability emerges when the scale of model parameters reaches a certain level. For instance, widely known LLMs typically possess billions of parameters. With such a powerful pre-trained model, a simple textual prompt can distill specific knowledge for downstream tasks. However, when it comes to prompting on graph tasks, the conditions become more intractable since pre-trained graph models have significantly fewer parameters, making it challenging to harness their full downstream adaptation potential.

\textbf{Intuitive Evaluation of Graph Prompt Learning}
In the NLP field, prompts are typically in a discrete textual format \cite{brown2020language, robinson2023leveraging}, allowing for intuitive understanding, comparison, and explanation. However, existing graph prompts are represented as learnable tokens \cite{sun2022gppt, fang2023universal, fang2022prompt, tan2023virtual, ma2023hetgpt, ge2023enhancing} or augmented graphs \cite{sun2023all, huang2023prodigy}. Such format poses challenges in intuitively understanding and interpreting graph prompts, as they lack a readable design. As a result, the effectiveness of prompts can only be evaluated based on downstream tasks, limiting efficient and comprehensive performance comparison among different kinds of graph prompts. Therefore, the development of a more intuitive graph prompt design with a readable format remains an open problem.

\textbf{More Downstream Applications}
Currently, graph prompt learning is primarily applied to node or graph classification tasks on open-source benchmarks \cite{hamilton2017inductive, xu2018how}. Although potential applications have been discussed in Section \ref{sec:app}, the real-world utilization of graph prompt learning remains limited. A notable example is its use in fraud detection within real-world transaction networks, addressing the issue of label scarcity \cite{wen2023voucher}. However, compared to the widespread application of prompting techniques in the NLP domain \cite{bran2023transformers,  robinson2023leveraging, zhang2023benchmarking}, the potential of graph prompt learning in diverse real-world applications requires further exploration.  Overcoming the main challenges of obtaining powerful domain-specific pre-trained graph models and designing suitable prompts for specific application scenarios that exhibit unique characteristics remains crucial.

\textbf{Transferable Prompt Designs} 
Existing studies on graph prompt learning \cite{liu2023graphprompt, sun2022gppt, gong2023prompt, fang2023universal, fang2022prompt, ge2023enhancing} typically focus on pre-training and prompt tuning using the same dataset, which limits the exploration of more transferable designs and empirical evaluation. Although PRODIGY \cite{huang2023prodigy} explores transferability within the same domain and All in One \cite{sun2023all} provides empirical results regarding transferability across tasks and domains, the investigation of prompt learning across diverse domains and tasks remains limited. Achieving transferability across diverse tasks and domains requires aligning the task space \cite{sun2023all}, semantic features \cite{zhu2023graphcontrol}, and structural patterns \cite{zhao2023graphglow}, which necessitates further theoretical work to provide insights guiding the development of transferable prompt designs.

\subsection{Future Directions}
With the above analysis on graph prompt, we summarize future directions as follows:

\textbf{Learning Knowledge from Large Graph Models (LGMs) like LLMs.} Currently, graph models are typically tailored to specific domains or tasks, limiting generalization abilities across broader domains or tasks \cite{liu2023graph, zhu2023graphcontrol}. Therefore, we are expecting the realization of large graph models (LGMs) as a universal tool to be intelligent enough to handle graph tasks across domains \cite{liu2023one, fatemi2023talk}. A significant step towards this direction has been taken by \citet{liu2023one}, who proposed a powerful model (OFA) capable of addressing classification tasks for graphs coming from various domains. However, their approach still relies on LLMs as a general template to distill specific domain knowledge, which we believe can be replaced with suitable graph prompting techniques. Just as textual prompt has been widely used to adapt LLM for diverse applications \cite{robinson2023leveraging, wang2023knowledge, brown2020language}, graph prompting techniques are promising to distill LGMs' knowledge specific for concrete downstream tasks. Applying the graph prompts to LGMs has the potential to revolutionize the field of deep graph learning, leveraging LGMs as a universal tool to tackle different tasks on graphs from diverse domains.

\textbf{Transferable Learning.} As we have discussed in Section \ref{sec:challenges}, current graph prompt learning methods \cite{sun2023all, sun2022gppt, huang2023prodigy, liu2023graphprompt, tan2023virtual, ge2023enhancing, shirkavand2023deep} are primarily limited to intra-dataset/domain settings, lacking the ability to transfer knowledge across different tasks or domains. While there have been efforts in graph transfer learning \cite{zhu2023graphcontrol} to realize domain adaptation, research specifically focused on transferable graph prompting techniques remains limited. To enable the transfer ability, the prompts should be designed to realize the alignment between different task spaces \cite{sun2023all}, reformulating different tasks on graphs into a uniform template. Besides, both structural \cite{zhao2023graphglow} and semantics alignment \cite{zhu2023graphcontrol} should be realized via suitable graph prompts to enable domain adaptation. It is promising to implement transferable graph prompts, realizing knowledge transfer across domains to extend the generalization and applicability of graph models.

\textbf{More Theoretical Foundation.} Despite the great success of graph prompt learning on various downstream tasks, they mostly draw on the successful experience of prompt tuning on NLP domain \cite{liu2022ptuning, brown2020language, gao2021making, tsimpoukelli2021multimodal, qin2021learning, shin2020autoprompt}. In other words, most existing graph prompt tuning methods are designed with intuition, and their performance gains are evaluated by empirical experiments. The lack of sufficient theoretical foundations behind the design has led to both performance bottlenecks and poor explainability. Therefore, we believe that building a solid theoretical foundation for graph prompt learning from a graph theory perspective and minimizing the gap between the theoretical foundation and empirical design is also a promising future direction.

\textbf{More Explainable and Understandable Design. }While existing graph prompt learning methods have demonstrated impressive results on various downstream tasks \cite{guo2023datacentric, wen2023voucher, yang2023empirical}, we still lack a clear understanding of what exactly is being learned from the prompts. The black-box learning mode of prompt vectors raises questions about their interpretability and whether we can establish meaningful correspondences between the input data and the prompted graph \cite{sun2023all}. These issues are crucial for understanding and interpreting prompts but are currently missing in most graph prompt research. To work towards trustworthy graph prompt learning, it is promising to explore the self-interpretability \cite{dai2021selfexplainable}
to enable intuitive explanations of graph prompts. By gaining insights into the learned prompt vectors and structures, we can enhance our understanding of the underlying mechanisms and improve the interpretability of graph prompts. This, in turn, can lead to more effective utilization of prompts for security- or privacy-related downstream tasks. 
\section{Conclusion}\label{sec:con}
In this survey, we explore the promising intersection between Artificial General Intelligence and graph data by graph prompt. Our unified framework has unveiled a structured understanding of graph prompts, dissecting them into tokens, token structures, and inserting patterns. This framework is a novel contribution, providing clarity and comprehensiveness for researchers and practitioners. By exploring the interplay between graph prompts and models, we've revealed fresh insights into the essence of graph prompts, highlighting their pivotal role in reshaping AI for graph data. With the development of ProG, a Python library, and a dedicated website, we've expanded the graph prompting ecosystem, enhancing collaboration and access to research, benchmark datasets, and code implementations. Our survey outlines a roadmap for the future. The challenges and future directions we've discussed serve as a beacon for the evolving field of graph prompting. With the above work, we hope our survey can push forward a new era of insights and applications in AGI family.
\section*{Acknowledgments}

The work was supported by grants from the Research Grant Council of the Hong Kong Special Administrative Region, China (Project No. CUHK 14217622), and CUHK Direct Grant No. 4055159. \dotuline{The first author, Dr. Xiangguo Sun, in particular, wants to thank his parents for their kind support during his tough period.} 

\noindent \textbf{Author Contributions:} 
\vspace*{-0.5\baselineskip} 
\begin{itemize}
    \item \textbf{Xiangguo Sun:} wrote section \ref{sec:intro}, section \ref{sec:method}, section \ref{subsec:why_prompt} (with Xixi), the main content of section \ref{sec:ptask}, section \ref{subsec:pmodal}, did proofreading of the whole paper, and proposed most of the original insights.
    \item \textbf{Jiawen Zhang:} wrote most of section \ref{sec:pre} (section \ref{subsec:notation}-\ref{subsec:his_prompt}), section \ref{sec:pretrain_method}, section \ref{sec:ptask}(collect, analyze papers and supplement insightful content), section \ref{sec:prog}, and were in charge of library development.
    \item \textbf{Xixi Wu:} wrote section \ref{subsec:why_prompt}, section \ref{sec:ptask} (collect, analyze papers and supplement insightful content), section \ref{sec:app}, most of section \ref{sec:dis} (challenges and future directions), collected and analyzed the references, and drew most of the figures.  
    \item \textbf{Hong Cheng,  Yun Xiong, and Jia Li:} were in charge of proofreading, and discussion, and contributed with many insightful opinions.
\end{itemize}

\normalem 
\bibliographystyle{IEEEtranSN}
\bibliography{zotero.bib}

\vspace{-10 mm}
\begin{IEEEbiography}[{\includegraphics[width=1in,height=1in,clip,keepaspectratio]{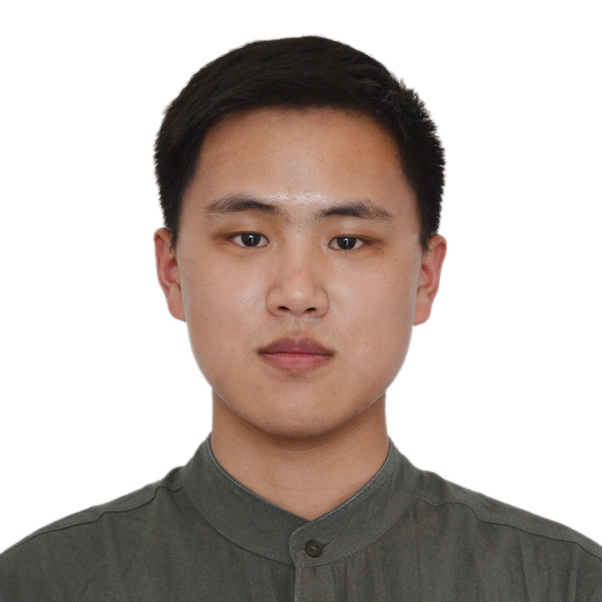}}]{Xiangguo Sun} is a postdoctoral research fellow at the Chinese University of Hong Kong. He was recognized as the "Social Computing Rising Star" in 2023 from CAAI. He studied at  Zhejiang Lab as a visiting researcher in 2022. In the same year, he received his Ph.D. from Southeast University and won the Distinguished Ph.D. Dissertation Award. During his Ph.D. study, he worked as a research intern at Microsoft Research Asia (MSRA) from Sep 2021 to Jan 2022 and won the ''Award of Excellence''. He studied as a joint Ph.D. student at The University of Queensland hosted by ARC Future Fellow Prof. Hongzhi Yin from Sep 2019 to Sep 2021. His research interests include social computing and network learning. He was the winner of the Best Research Paper Award at KDD'23, which is the first time for Mainland and Hong Kong of China. 
He has published 11 CORE A*, 9 CCF A, and 13 SCI (including 6 IEEE Trans), some of which appear in SIGKDD, VLDB, The Web Conference (WWW), TKDE, TOIS, WSDM, TNNLS, CIKM, etc.
\end{IEEEbiography}
\vspace{-10 mm}

\begin{IEEEbiography}[{\includegraphics[width=1in,height=1in,clip,keepaspectratio]{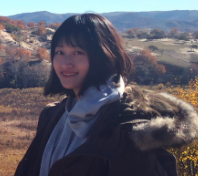}}]{Jiawen Zhang} is a Ph.D. student in Data Science and Analytics at HKUST (Guangzhou) under the supervision of Prof. Jia Li. She received her M.Eng. degree in Computer Technology from the Chinese Academy of Sciences. She worked as a research intern at the Machine Learning Group in Microsoft Research Asia. Her research interests include data mining and time series modeling.
\end{IEEEbiography}
\vspace{-13 mm}

\begin{IEEEbiography}[{\includegraphics[width=1in,height=1in,clip,keepaspectratio]{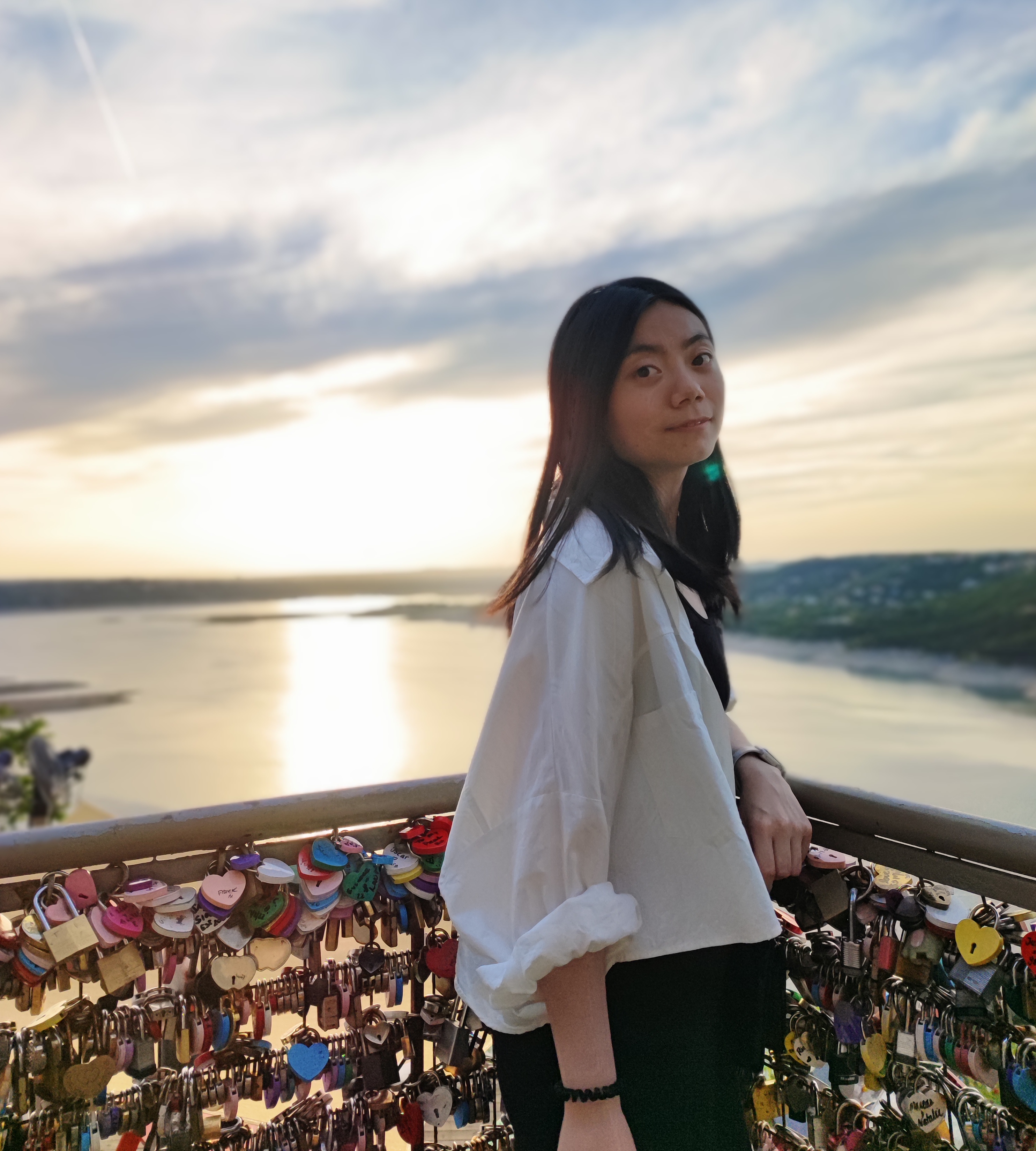}}]{Xixi Wu} is a final-year master student at School of Computer Science, Fudan University. She obtained her B.S. in Computer Science from Fudan University in 2021. Her research interests include Deep Graph Learning, Data Mining, and Recommender Systems. She has published multiple first-authored papers at KDD/WWW/CIKM.
\end{IEEEbiography}
\vspace{-13 mm}

\begin{IEEEbiography}[{\includegraphics[width=1in,height=1in,clip,keepaspectratio]{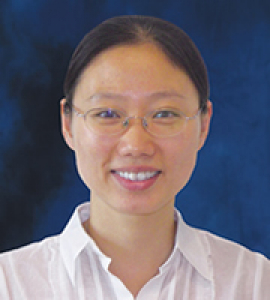}}]{Hong Cheng} is a Professor in the Department of Systems Engineering and Engineering Management, Chinese University of Hong Kong. She received the Ph.D. degree from the University of Illinois at Urbana-Champaign in 2008. Her research interests include data mining, database systems, and machine learning. She received research paper awards at ICDE'07, SIGKDD'06, and SIGKDD'05, and the certificate of recognition for the 2009 SIGKDD Doctoral Dissertation 
 Award. She received the 2010 Vice-Chancellor's Exemplary Teaching Award at the Chinese University of Hong Kong.
\end{IEEEbiography}
\vspace{-10 mm}

\begin{IEEEbiography}[{\includegraphics[width=1in,height=1in, clip,keepaspectratio]{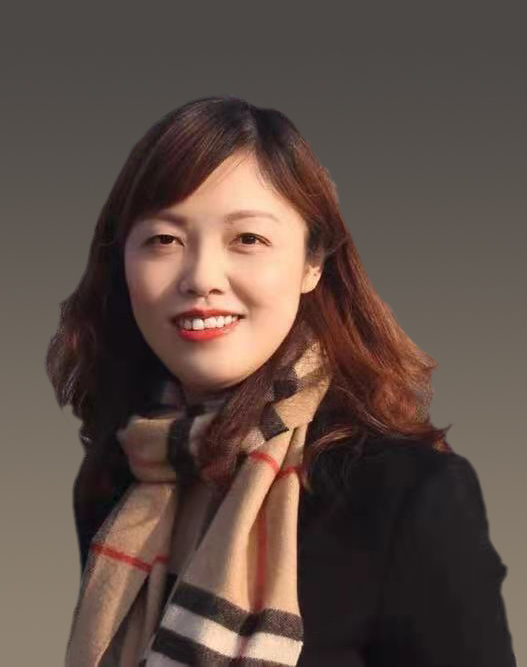}}]{Yun Xiong} is a Professor in Shanghai Key Laboratory of Data Science, School of Computer Science, Fudan University. She obtained her Ph.D. degree from Fudan University in 2008. Her research interests include big data mining and data science. Her research achievements include the publication of more than 50 papers in internationally top journals and conferences in the field of Data Mining, including TKDE, KDD, WWW, AAAI, etc.
\end{IEEEbiography}
\vspace{-13 mm}

\begin{IEEEbiography}[{\includegraphics[width=1in,height=1in,clip,keepaspectratio]{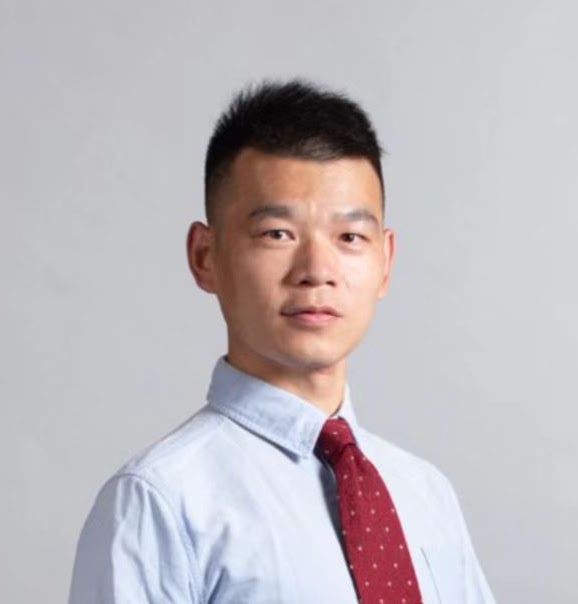}}]{Jia Li} is an assistant professor in HKUST (Guangzhou). He received Ph.D. degree at The Chinese University of Hong Kong in 2021. Before that, he worked as a full-time data mining engineer at Tencent from 2014 to 2017, and research intern at Google AI (Mountain View) in 2020. His research interests include graph learning and data mining. Some of his work has been published in TPAMI, ICML, NeurIPS, WWW, KDD, etc.
\end{IEEEbiography}

\vfill

\end{document}